%% file: ex_arxiv.tex
\documentclass[letterpaper,11pt]{article}
\usepackage[margin=1in,letterpaper]{geometry} 
\input{math_commands.tex}

\usepackage{hyperref}
\usepackage{url}

\usepackage{bm}
\usepackage{subfigure}
\usepackage{amsmath}
\usepackage{amsthm}
\usepackage{amssymb}
\usepackage{amsfonts}
\usepackage{amsopn}
\usepackage{verbatim}
\usepackage{algorithmic,algorithm}
\usepackage{color}
\usepackage{enumitem}
\usepackage{stfloats}
\usepackage{float}
\usepackage{multirow}
\usepackage{siunitx}
\usepackage{graphicx}
\usepackage{mathrsfs}
\usepackage{thmtools,thm-restate}
\usepackage{threeparttable}

\usepackage{amsthm}
\usepackage{bm}
\usepackage{color}
\usepackage{enumitem}
\usepackage{stfloats}
\usepackage{float}
\usepackage{multirow}
\usepackage{siunitx}
\usepackage{amsfonts}
\usepackage{amsmath}
\usepackage{amssymb}
\usepackage{mathtools}
\usepackage{amsthm}
\usepackage{wrapfig}

\usepackage{url}
\usepackage{threeparttable}
\newtheorem{theorem}{Theorem}

\newtheorem{lemma}[theorem]{Lemma}

\title{Decentralized Entropic Optimal Transport for Distributed Distribution Comparison}

\author{Xiangfeng Wang$^1$\quad Hongteng Xu$^2$\quad Moyi Yang$^1$\thanks{The authors have equal contributions and are listed in alphabetical order of their last names.} \\
$^1$School of Computer Science and Technology, East China Normal University \\
$^2$Gaoling School of Artificial Intelligence, Renmin University of China \\
\texttt{xfwang@cs.ecnu.edu.cn}\quad \texttt{hongtengxu@ruc.edu.cn}\quad
\texttt{winnie\_yang@stu.ecnu.edu.cn} \\
}

\begin{document}

\maketitle

\begin{abstract}
Distributed distribution comparison aims to measure the distance between the distributions whose data are scattered across different agents in a distributed system and cannot even be shared directly among the agents. 
In this study, we propose a novel decentralized entropic optimal transport (DEOT) method, which provides a \textit{communication-efficient} and \textit{privacy-preserving} solution to this problem with theoretical guarantees. 
In particular, we design a mini-batch randomized block-coordinate descent (MRBCD) scheme to optimize the DEOT distance in its dual form.
The dual variables are scattered across different agents and updated locally and iteratively with limited communications among partial agents. 
The kernel matrix involved in the gradients of the dual variables is estimated by a decentralized kernel approximation method, in which each agent only needs to approximate and store a sub-kernel matrix by one-shot communication and without sharing raw data. 
Besides computing entropic Wasserstein distance, we show that the proposed MRBCD scheme and kernel approximation method also apply to entropic Gromov-Wasserstein distance. 
We analyze our method's communication complexity and, under mild assumptions, provide a theoretical bound for the approximation error caused by the convergence error, the estimated kernel, and the mismatch between the storage and communication protocols. 
In addition, we discuss the trade-off between the precision of the EOT distance and the strength of privacy protection when implementing our method.
Experiments on synthetic data and real-world distributed domain adaptation tasks demonstrate the effectiveness of our method.
\end{abstract}



\input{sections-arxiv/intro.tex}

\input{sections-arxiv/related.tex}

\input{sections-arxiv/method.tex}

\input{sections-arxiv/extension.tex}

\input{sections-arxiv/exp.tex}

\section{Conclusion}\label{sec:conclusion}
In this study, we proposed a decentralized mini-batch randomized block-coordinate descent scheme to approximate the EOT distance in a decentralized scenario and analyzed the approximation error in theory.
Our method is communicate-efficient and privacy-preserving and can be extended to compute EGW distance, which provides a potential solution to various distributed distribution comparison tasks. 
In the future, we plan to accelerate our method based on the importance sparsification and extend it to more challenging scenarios, e.g., approximating the DEOT for continuous distributions and achieving decentralized fused Gromov-Wasserstein distance~\cite{titouan2019optimal}.

\bibliography{refs}
\bibliographystyle{IEEEtran}

\end{document}

%% file: math_commands.tex
\usepackage{amsmath,amsfonts,bm}

\def\eqref#1{equation~\ref{#1}}

\def\1{\bm{1}}

\DeclareMathAlphabet{\mathsfit}{\encodingdefault}{\sfdefault}{m}{sl}
\SetMathAlphabet{\mathsfit}{bold}{\encodingdefault}{\sfdefault}{bx}{n}



%% file: sections-arxiv/intro.tex
\section{Introduction}\label{sec:intro}
Distribution comparison plays a central role in many machine learning problems, such as data clustering~\cite{hammouda2000comparative}, generative modeling~\cite{bond2021deep,mattei2019miwae}, domain adaptation~\cite{farahani2021brief,ganin2015unsupervised}, etc. 
As a valid metric for distributions, optimal transport (OT) distance~\cite{villani2009optimal} provides a powerful solution to this task.
Mathematically, given two distributions in a compact space $\mathcal{X}$, denoted as $\mu,\gamma\in\mathcal{P}(\mathcal{X})$, the OT distance between them corresponds to the minimum expectation of the costs defined on their sample pairs, in which the optimal distribution of the sample pairs, or called optimal coupling, takes $\mu$ and $\gamma$ as its marginals, respectively.
The Kantorovich formulation of the OT distance is defined as
\begin{eqnarray}\label{eq:ot}
\begin{aligned}
    W(\mu, \gamma) \stackrel{\text {def.}}{=} \inf_{\pi \in \Pi(\mu, \gamma)} \int_{\mathcal{X}\times\mathcal{X}} c(\bm{x}, \bm{y})\pi(\bm{x}, \bm{y}) \mathrm{d}\bm{x}\mathrm{d}\bm{y}, 
\end{aligned}
\end{eqnarray}
where $c:\mathcal{X}\times\mathcal{X}\mapsto \mathbb{R}$ denotes a continuous cost function, the distribution on the product space $\mathcal{X} \times \mathcal{X}$ is denoted as the coupling $\pi$, and $\Pi(\mu, \gamma) =\{ \pi \in \mathcal{P}(\mathcal{X} \times \mathcal{X}) \ \big|\ \forall({\mathcal{A}}, {\mathcal{B}}) \subset \mathcal{X} \times \mathcal{X}, \pi({\mathcal{A}} \times \mathcal{X})=\mu({\mathcal{A}}), \pi(\mathcal{X} \times {\mathcal{B}})=\gamma({\mathcal{B}})\}$ denotes the marginal constraints of the coupling. 
When the cost function $c$ is a norm-induced distance metric, the OT distance becomes the so-called Wasserstein distance~\cite{villani2009optimal}.
The OT distance is applicable even if the supports of the two distributions are non-overlapped. 
Therefore, leveraging it to fit a model distribution to the data distribution~\cite{arjovsky2017wasserstein,deshpande2018generative} or transferring a source distribution to a target one~\cite{courty2014domain,damodaran2018deepjdot} often leads to encouraging performance.

In practice, the optimal transport distance in~\eqref{eq:ot} is often implemented with an entropic regularizer~\cite{blondel2018smooth,cuturi2013sinkhorn}, which leads to a strictly-convex optimization problem called entropic optimal transport (EOT):
\begin{eqnarray}\label{eq:eot}
\begin{aligned}
    W_{\varepsilon}(\mu, \gamma) \stackrel{\text {def.}}{=} \inf_{\pi \in \Pi(\mu, \gamma)} \int_{\mathcal{X}\times \mathcal{X}} c(\bm{x}, \bm{y})\pi(\bm{x}, \bm{y}) \mathrm{d}\bm{x}\mathrm{d}\bm{y} - \varepsilon H(\pi), 
\end{aligned}
\end{eqnarray}
where $H(\pi)=-\int_{\mathcal{X} \times \mathcal{X}}\pi(\bm{x},\bm{y})\log (\pi(\bm{x},\bm{y}) -1 )\ \mathrm{d}\bm{x}\mathrm{d}\bm{y}$ denotes the entropy of $\pi$, whose significance is controlled by $\varepsilon>0$.
Given the samples of $\mu$ and $\gamma$, this problem can be solved efficiently using various alternate minimization schemes, e.g., the Sinkhorn-scaling algorithm in~\cite{cuturi2013sinkhorn} and its stochastic version in~\cite{altschuler2017near}, the Bregman alternating direction method of multipliers (BADMM) in~\cite{wang2014bregman}, the primal-dual method in~\cite{blondel2018smooth}, and the block-coordinate descent method in~\cite{genevay2016stochastic}. 

\begin{figure}[t]
    \centering
    \includegraphics[width=1.0\linewidth]{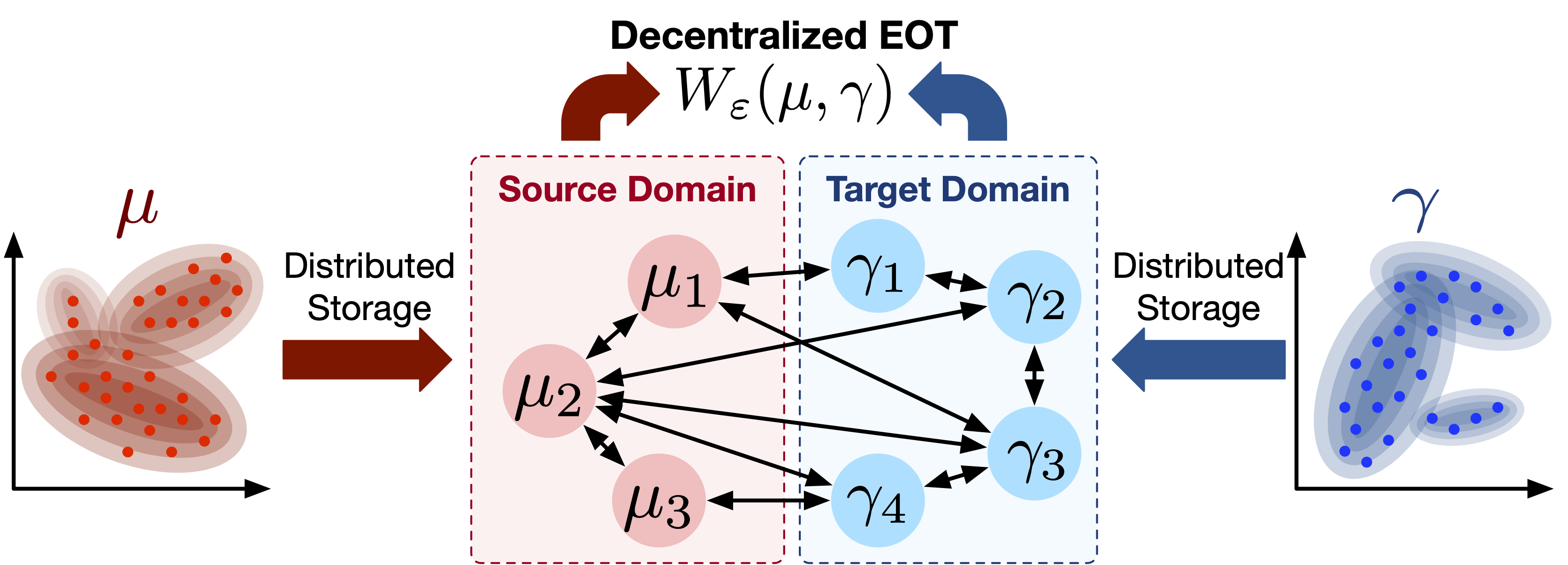}
    \caption{An illustration of the distributed distribution comparison task and the corresponding decentralized entropic optimal transport problem. 
    To emphasis, the data in two domains obey the distribution $\mu$ and $\gamma$, respectively, but are scattered among different agents.
    Each agent merely contains a part of samples, and the local distribution is denoted as $\mu_i$ (or $\gamma_j$). 
    Communication between different agents is allowed, but it obeys the communication protocol relevant to the network topology. 
    Additionally, in the privacy-preserving setting, the communication of raw data is forbidden.}
    \label{fig:diagram}
\end{figure}

The above algorithms are generally centralized --- a central server is required to collect all samples and their pairwise costs and then solve~\eqref{eq:eot} accordingly. 
However, in many real-world applications, the samples of a distribution are often large-scale and scattered across different agents in a distributed system (i.e., each agent only has limited storage and computation power, and none works as a central server). 
In addition, sharing raw data directly can be forbidden in this system because of privacy and security requirements. 
As illustrated in Fig.~\ref{fig:diagram}, such a scenario leads to a challenging \textit{distributed distribution comparison} task, in which each agent can access neither the whole sample sets nor the complete cost matrix $\bm{C}=[c(\bm{x},\bm{y})]$.
Therefore, a decentralized method is required to solve the EOT problem efficiently with limited, even privacy-preserving, communications.

We propose a novel decentralized entropic optimal transport (DEOT) method in this study, which provides an efficient and theoretically-guaranteed solution to distributed distribution comparison. 
In particular, our method considers the dual form of the EOT problem, in which the dual objective involves a kernel associated with the cost function, and the dual variables are scattered among the agents. 
We approximate a sub-kernel matrix for each agent via the decentralized kernel approximation method in~\cite{khanduri2021decentralized}, which only requires one-shot communication and avoids sharing raw data among the agents. 
Based on the approximated kernel, the dual variables are optimized in a mini-batch randomized block-coordinate descent (MRBCD) scheme~\cite{zhao2014accelerated}. 
Each agent stores and updates the dual variables corresponding to its local data. 
The dual variables' gradients are computed based on partial dual variables (rather than raw data) from some randomly-selected agents. 
The convergence of the method is guaranteed in theory. 
Besides solving the classic EOT problem in~\eqref{eq:eot}, our DEOT method is also applicable for entropic Gromov-Wasserstein (EGW) distance~\cite{rioux2023entropic,zhang2022gromov} under slight modifications. 

The proposed method is communication-efficient in high-dimensional scenarios because the communication cost is independent of the data dimension.
In addition, without sharing raw data, this method can achieve a trade-off between the precision of the EOT distance and the strength of privacy protection.
Moreover, as our main theoretical contribution, under mild assumptions, we make the first attempt to provide an error bound of the DEOT distance caused by the algorithmic convergence error, the kernel approximation error, and the mismatch between the distributed system's storage and communication protocols.

Experiments on synthetic data verify the effectiveness of our DEOT method and its robustness to hyperparameter settings and communication protocols.
Furthermore, we test our DEOT method in real-world distributed domain adaptation tasks, demonstrating its usefulness in practice.

%% file: sections-arxiv/related.tex
\section{Related Work}\label{sec:related}
\subsection{Entropic Optimal Transport Methods}
Entropic optimal transport distance can be computed by the Sinkhorn-scaling algorithm in~\cite{benamou2015iterative,cuturi2013sinkhorn,sinkhorn1967concerning} (or its logarithmic variant~\cite{chizat2018scaling,schmitzer2019stabilized} for improving numerical stability). 
Following the Sinkhorn-scaling algorithm, the method in~\cite{xie2020fast} computes the OT distance via an inexact proximal point algorithm, which is equivalent to solving an EOT problem with a temporally-decayed entropic regularizer. 
The Greenkhorn algorithm in~\cite{altschuler2017near} works as a stochastic Sinkhorn-scaling algorithm with a significant improvement in computational efficiency. 
Besides the Sinkhorn-scaling algorithm, some other efficient algorithms are developed, e.g., BADMM~\cite{wang2014bregman,ye2017fast}, smoothed semi-dual algorithm~\cite{blondel2018smooth}, and conditional gradient (CG) algorithm~\cite{titouan2019optimal}.
The work in~\cite{genevay2016stochastic,seguy2018largescale} introduce stochastic optimization mechanisms into the large-scale optimal transport problem.

The above methods mainly focus on solving the EOT problem in centralized scenarios.
The decentralized EOT problem is seldom studied.
Recently, the work in~\cite{hughes2021fair,zhang2019consensus} proposes some ADMM-based decentralized algorithms for distributed resource allocation tasks.
The formulation of their tasks is relevant to an optimal transport problem rather than an EOT problem.
Moreover, unlike our work, their methods neither apply any privacy-preserving mechanism nor consider the mismatch between the distributed system's storage and communication protocols.

Note that the decentralized EOT problem differs from the well-known distributed and decentralized Wasserstein barycenter problems in~\cite{dvurechenskii2018decentralize,staib2017parallel,uribe2018distributed}.
In these barycenter problems, the samples of a distribution are still stored in a single agent so that the (entropic) optimal transport distance between the distribution and the barycenter can still be computed in a centralized manner.
On the contrary, in our study, each agent can only access partial samples of a distribution and cannot share them with other agents. 
Hence, we need a decentralized algorithm to compute the EOT distance.

\subsection{Distributed and Decentralized Optimization}
Distributed and decentralized optimization methods can be broadly divided into primal and primal-dual strategies~\cite{duchi2011dual,nedic2009distributed}. 
The primal strategy usually refers to gradient-based methods such as decentralized gradient descent (DGD)~\cite{lobel2010distributed,yuan2016convergence}, EXTRA~\cite{shi2015extra}, etc.
For large-scale optimization tasks, the stochastic gradient technique is often applied, e.g., the decentralized stochastic gradient descent method (D-SGD)~\cite{agarwal2011distributed} is generalized from the DGD, with a significant improvement in computational efficiency. 
The primal-dual strategy further introduces dual variables to design distributed optimization methods, which incorporates distributed dual decomposition~\cite{terelius2011decentralized}, ADMM~\cite{ChangHW15,shi2014linear}, etc.
More discussions can refer to a series of survey papers~\cite{assran2020advances,ChangHWZL20,Nedic20}.

Besides computational efficiency, communication efficiency is also required for distributed and decentralized optimization. 
The most intuitive way to increase the communication efficiency is to reduce the number of agents involved in communications~\cite{smith2018cocoa}, e.g., the random node selection scheme in~\cite{arablouei2015analysis,yin2018communication} and the importance sampling scheme in~\cite{chen2018lag,liu2019communication}. 
To reduce the bandwidth, we often compress the information for each communication by sparsification~\cite{basu2019qsparse,stich2018sparsified,tang2020communication} or quantization~\cite{alistarh2017qsgd,lu2020moniqua,zhang2019quantized,zhu2016quantized}.

Recently, the distributed and decentralized optimization techniques have been utilized for distributed deep learning~\cite{tang2020communication}, distributed edge AI system~\cite{shi2020communication}, federated learning~\cite{chen2021communication}, etc. 
These applications often consider data privacy issues when comparing distributions.
Typically, the classic differential privacy strategy~\cite{dwork2006differential} adds random noise to data before communication, which can protect data privacy with theoretical guarantees~\cite{friedman2010data,wasserman2010statistical} and has been widely used in federated learning scenarios~\cite{truex2020ldp,wei2020federated}.
Recently, the compressive sensing technique~\cite{candes2006stable} is also applicable for privacy protection~\cite{testa2019compressed,wang2014privacy,xiong2022compressive} --- instead of sharing raw data, we can apply random projection to the data and share the projection results. 
Additionally, for some special kinds of data, some sophisticated privacy-preserving methods can be applied, e.g., the decentralized kernel approximation method in~\cite{khanduri2021decentralized} for the Gram matrix of the kernel function. 

Motivated by the above methods, we develop the decentralized EOT method with privacy preservation and communication efficiency.
We demonstrate the rationality of the proposed method in theory and apply it to various privacy-preserving distributed distribution comparison tasks.

%% file: sections-arxiv/method.tex
\section{Proposed Decentralized Entropic Optimal Transport}\label{sec:method}

\subsection{Dual Formulation of Decentralized EOT}
Suppose that there are $I$ agents in the source domain storing the samples of $\mu$ and $J$ agents in the target domain storing the samples of $\gamma$, as illustrated in Fig.~\ref{fig:diagram}. 
The distribution of the samples in the $i$-th source agent (the $j$-th target agent) is denoted as $\mu_i$ ($\gamma_j$). 
Accordingly, the storage of the samples in the agents can be captured by the following hierarchical model:
\begin{eqnarray}\label{eq:data}
\begin{aligned}
    \text{Agent selection:}&\quad i \sim \bm{p},\quad j\sim \bm{q};\\
    \text{Sample assignment:}&\quad \bm{X}_i=\{\bm{x}_n^{(i)}\}_{n=1}^{N_i}\sim \mu_i,\quad \bm{Y}_j=\{\bm{y}_m^{(j)}\}_{m=1}^{M_j}\sim \gamma_j;
\end{aligned}
\end{eqnarray}where $\bm{p}=\{p_i\}_{i=1}^{I}\in\Delta^{I-1}$ and $\bm{q}=\{q_j\}_{j=1}^{J}\in\Delta^{J-1}$ indicate the distribution of the source agents and that of the target agents, respectively.
$p_i$ ($q_j$) represents the probability of selecting the source agent $i$ (the target agent $j$) to store the corresponding data.
$\bm{X}_i=\{\bm{x}_n^{(i)}\}_{n=1}^{N_i}$ and $\mu_i$ denote the samples stored in the agent $i$ and the corresponding distribution, respectively.
$\bm{Y}_j=\{\bm{y}_m^{(j)}\}_{m=1}^{M_j}$ and $\gamma_j$ are denoted in the same way. 
Obviously, we have $\mu=\sum_i p_i \mu_i$, $\gamma =\sum_j q_j \gamma_j$, $\bm{X}=\{\bm{x}_n\}_{n=1}^{N}=\cup_{i=1}^{I} \bm{X}_i$, and $\bm{Y}=\{\bm{y}_m\}_{m=1}^{M}=\cup_{j=1}^{J} \bm{Y}_j$. 

Taking the Fenchel dual form of the EOT distance~\cite{peyre2019computational} into account, we rewrite the EOT distance in~\eqref{eq:eot} as follows:
\begin{eqnarray}\label{eq:dual-eot}
\begin{aligned}
    W_{\varepsilon}(\mu, \gamma)
    =& \sideset{}{}\sup_{u, v \in \mathcal{C}_{\mathcal{X}}} \int_{\mathcal{X}} u(\bm{x})\mu(\bm{x}) \mathrm{d}\bm{x}  + \int_{\mathcal{X}} v(\bm{y})\gamma(\bm{y}) \mathrm{d}\bm{y} \\
    &\qquad -\varepsilon \int_{\mathcal{X}\times\mathcal{X}} e^{\frac{u(\bm{x})+v(\bm{y})-c(\bm{x}, \bm{y})}{\varepsilon}}\mu(\bm{x})\gamma(\bm{y}) \mathrm{d}\bm{x}\mathrm{d}\bm{y} \\
    =&\sideset{}{}\sup_{u, v \in \mathcal{C}_{\mathcal{X}}} \mathbb{E}_{\bm{x}\sim \mu,\bm{y}\sim\gamma}f_{\varepsilon}(u,v;\kappa(\bm{x},\bm{y}))\\
    =&\sideset{}{}\sup_{u, v \in \mathcal{C}_{\mathcal{X}}} \mathbb{E}_{(i,j)\sim \bm{p}\bm{q}^{\top}}\mathbb{E}_{\bm{x}\sim \mu_i,\bm{y}\sim\gamma_j}f_{\varepsilon}(u,v;\kappa(\bm{x},\bm{y})).
\end{aligned} 
\end{eqnarray}
Here, $\mathcal{C}_{\mathcal{X}}$ represents the set of continuous functions defined in $\mathcal{X}$, $u,v\in \mathcal{C}_{\mathcal{X}}$ denote the dual functions, which are also called Kantorovich potentials, and 
\begin{eqnarray}\label{eq:f-varepsilon}
\begin{aligned}
    f_{\varepsilon}(u,v;\kappa(\bm{x},\bm{y}))=u(\bm{x})+v(\bm{y})-\varepsilon e^{\frac{u(\bm{x}) + v(\bm{y})}{\varepsilon}}\underbrace{e^{-\frac{c(\bm{x}, \bm{y})}{\varepsilon}}}_{\kappa(\bm{x},\bm{y})},
\end{aligned}
\end{eqnarray}
where $\kappa(\bm{x},\bm{y})$ is a kernel function associated with the cost $c(\bm{x},\bm{y})$. 
The second equation in~\eqref{eq:dual-eot} indicates that the EOT problem can be modeled as an unconstrained expectation maximization problem with respect to $u$ and $v$~\cite{genevay2016stochastic}.
The third equation in~\eqref{eq:dual-eot} is based on the hierarchical model in~\eqref{eq:data}, which leads to the proposed DEOT problem. 
Note that, as shown in~\eqref{eq:dual-eot}, we sample source and target agents independently from $\bm{p}$ and $\bm{q}$, which is equivalent to sampling the agent pairs from the distribution $\bm{pq}^{\top}=[p_iq_j]$.
In the following content, we define $\bm{pq}^{\top}$ as the \textit{storage protocol} of the distributed system. 

\subsection{DEOT with A Communication Protocol}

As shown in~\eqref{eq:dual-eot}, computing the DEOT distance requires us to sample agent pairs based on the storage protocol. 
In practice, however, the sampling of the agent pairs is determined by the \textit{communication protocol} rather than the storage protocol of the distributed system. 
Here, we define the communication protocol as the distribution of the communicable agent pairs, denoted as $\bm{E}=[e_{ij}] \in \{\bm{E}\in\mathbb{R}_+^{I\times J}\, |\, {\bm{1}}_I^{\top}\bm{E} {\bm{1}}_J=1\}$. 
Generally, the communication protocol can be mismatched with the storage protocol.
For example, some systems do not allow multi-step routes and/or restrict the communication between the agents to be directed, which may cause $\bm{E}\neq \bm{pq}^{\top}$. 
As a result, we actually approximate $W_{\varepsilon}(\mu, \gamma)$ by the following surrogate:
\begin{equation}\label{eq:dual-eot2}
\begin{aligned}
    \widetilde{W}_{\varepsilon}(\mu, \gamma)
    \stackrel{\text {def.}}{=}\sideset{}{}\sup_{u, v \in \mathcal{C}_{\mathcal{X}}} \mathbb{E}_{(i, j)\sim \bm{E}}\mathbb{E}_{\bm{x}\sim \mu_i,\bm{y}\sim\gamma_j}f_{\varepsilon}(u,v;\kappa(\bm{x},\bm{y})).
\end{aligned}
\end{equation}
Obviously, $\widetilde{W}_{\varepsilon}(\mu, \gamma)=W_{\varepsilon}(\mu, \gamma)$ when $\bm{E}=\bm{pq}^{\top}$. 
In a distributed system built on a connected network and with a known storage protocol, we can first select a source agent based on $\bm{p}$ and then select a target agent based on $\bm{q}$ (so that $\bm{E}=\bm{pq}^{\top}$). 
In more general settings, we need to adjust the communication protocol, matching it with the storage protocol as much as possible. 


\subsection{Mini-Batch Randomized Block-Coordinate Descent}

Given the samples of $\mu$ and $\gamma$, i.e., $\{\bm{x}_n\}_{n=1}^{N}\sim\mu$ and $\{\bm{y}_m\}_{m=1}^{M}\sim\gamma$, the problem in~\eqref{eq:dual-eot2} becomes
\begin{eqnarray}\label{eq:sample-eot}
\begin{aligned}
    \max_{\substack{\bm{u}=\{\bm{u}^{(i)}\}_{i=1}^{I} \in \mathbb{R}^{N}\\ \bm{v}=\{\bm{v}^{(j)}\}_{j=1}^{J} \in \mathbb{R}^{M}}}
    \overbrace{
    \sideset{}{}\sum_{i=1}^{I}\sideset{}{}\sum_{j=1}^{J} \frac{e_{ij}}{N_iM_j} \underbrace{f_{\varepsilon}( \bm{u}^{(i)}, \bm{v}^{(j)};\bm{K}_{ij})}_{f_{\varepsilon}^{(i,j)}}}^{F_{\varepsilon}(\bm{u},\bm{v};\bm{K},\bm{E})},
\end{aligned}
\end{eqnarray}
where the dual functions $u$ and $v$ become the dual variables $\bm{u} \in \mathbb{R}^{N}$ and $\bm{v} \in \mathbb{R}^{M}$, respectively. 
The dual objective $F_{\varepsilon}(\bm{u},\bm{v};\bm{K},\bm{E})$ takes the kernel matrix $\bm{K}=[\kappa(\bm{x}_n,\bm{y}_m)]\in \mathbb{R}^{N\times M}$ and the communication protocol $\bm{E}$ as its hyperparameters.
The dual objective is decomposable - for the agent pair $(i,j)$, we have a local objective, i.e., $f_{\varepsilon}^{(i,j)}=f_{\varepsilon}( \bm{u}^{(i)}, \bm{v}^{(j)};\bm{K}_{ij})$, as follows,
\begin{eqnarray}\label{eq:local}
\begin{aligned}
    f_{\varepsilon}^{(i,j)}&=f_{\varepsilon}( \bm{u}^{(i)}, \bm{v}^{(j)};\bm{K}_{ij})\\
    &=\sum_{n=1}^{N_i}\sum_{m=1}^{M_j}f_{\varepsilon}(u_n^{(i)}, v_m^{(j)};\kappa(\bm{x}_n^{(i)}, \bm{y}_m^{(j)}))\\
    &=u_n^{(i)}+v_m^{(j)}-\varepsilon\exp\Bigl(\frac{u_n^{(i)}+v_m^{(j)}}{\varepsilon}\Bigr)\kappa(\bm{x}_n^{(i)}, \bm{y}_m^{(j)}),
\end{aligned}
\end{eqnarray}
in which $\bm{K}_{ij}=[\kappa(\bm{x}_n^{(i)}, \bm{y}_m^{(j)})]\in \mathbb{R}^{N_i\times M_j}$ is a block of $\bm{K}$.
Here, each local objective only involves a part of dual variables that correspond to the local samples stored in the agents, i.e., the $\bm{u}^{(i)}=[u_n^{(i)}]\in\mathbb{R}^{N_i}$ in $f_{\varepsilon}^{(i,j)}$ corresponds to the samples $\{\bm{x}_n^{(i)}\}_{n=1}^{N_i}$ in the source agent $i$. 
As a result, the dual variables can be scattered across different agents, and accordingly, the gradient of $\bm{u}^{(i)}$ can be formulated as follows:
\begin{eqnarray}\label{eq:grads_f}
\begin{aligned}
    \nabla_{\bm{u}^{(i)}}F_{\varepsilon}(\bm{u},\bm{v};\bm{K},\bm{E})&=\sum_{j=1}^{J} \frac{e_{ij}}{N_iM_j}\nabla_{\bm{u}^{(i)}}f_{\varepsilon}^{(i,j)}
    \\
    &=\sum_{j=1}^{J}\frac{e_{ij}}{N_iM_j}
\begin{bmatrix}
\sum_{m=1}^{M_j}1-e^{\frac{u_{1}^{(i)} + v_m^{(j)}}{\varepsilon}}\kappa(\bm{x}_1^{(i)}, \bm{y}_m^{(j)})\\
\vdots\\
\sum_{m=1}^{M_j}1-e^{\frac{u_{N_i}^{(i)} + v_m^{(j)}}{\varepsilon}}\kappa(\bm{x}_{N_i}^{(i)}, \bm{y}_m^{(j)})
\end{bmatrix}.
\end{aligned}
\end{eqnarray}

\begin{algorithm}[t]
\caption{MRBCD for $\max_{\bm{u},\bm{v}}F_{\varepsilon}(\bm{u},\bm{v};\bm{K},\bm{E})$}\label{Algo-MRBCD}
\begin{algorithmic}[1] 
\STATE \textbf{One-step data communication:}\hfill \textcolor{red}{$\mathcal{O}((IM+JN)D)$}
\STATE For each source agent $i$ and target agent $j$, construct $\{\bm{K}_{ij}\}_{j=1}^{J}$ and $\{\bm{K}_{ij}\}_{i=1}^{I}$ based on received data.
\STATE \textbf{Update dual variables:}\hfill \textcolor{red}{$\mathcal{O}(TL(\frac{N}{I}+\frac{M}{J}))$}
\STATE Initialize $\bm{u}^{(i)}=\bm{0}$ and $\bm{v}^{(j)}=\bm{0}$ randomly for $i=1,...,I$ and $j=1,...,J$.
\FOR{$t=0,1,\cdots,T$}
\STATE Set the learning rate $\eta_t=\frac{\eta}{\sqrt{t+1}}$.
\FOR{An agent pair $(i,j)\sim \bm{E}$}
\STATE Select $L$ target agents $\mathcal{J}_L\sim\frac{1}{\|\bm{E}_{i, :}\|_1}\bm{E}_{i, :}$. Send $\{\bm{v}^{(j),t}\}_{j\in\mathcal{J}_L}$ to the source agent $i$.
\STATE $\bm{u}^{(i),t+1}\leftarrow \bm{u}^{(i),t} + \eta_t\sum_{j\in\mathcal{J}_L}\nabla_{\bm{u}^{(i)}}{f}_{\varepsilon}^{(i,j),t}$
\STATE Select $L$ source agents $\mathcal{I}_L\sim\frac{1}{\|\bm{E}_{:,j}\|_1}\bm{E}_{:,j}$. Send $\{\bm{u}^{(i),t}\}_{i\in\mathcal{I}_L}$ to the target agent $j$
\STATE $\bm{v}^{(j),t+1}\leftarrow \bm{v}^{(j),t} + \eta_t\sum_{i\in\mathcal{I}_L}\nabla_{\bm{v}^{(j)}}{f}_{\varepsilon}^{(i,j),t}$
\ENDFOR
\ENDFOR
\STATE\textbf{Compute EOT distance and broadcast it:}\hfill \textcolor{red}{$\mathcal{O}(IJ)$}
\STATE For an arbitrary source agent $i$, receive the optimal dual objectives $\{\{{f}_{\varepsilon}^{(i',j)}\}_{j=1}^{J}\}_{i'\neq i}$ from the remaining agents in the source domain.
\STATE Compute $\widetilde{W}_{\varepsilon}(\mu,\gamma)$ in the source agent $i$ and broadcast it to all other agents.
\end{algorithmic}
\end{algorithm}

We propose a mini-batch randomized block coordinate descent (MRBCD) scheme, computing the gradient of $f_{\varepsilon}^{(i,j)}$ based on a batch of agents and optimizing the decentralized EOT problem iteratively. 
Take a source agent $i$ as an example. 
In the $t$-th iteration, the agent $i$ receives the dual variables $\{\bm{v}^{(j),t}\}_{j\in\mathcal{J}_L}$ from $L$ target agents, where $\mathcal{J}_L\subset\{1,...,J\}$ denotes the set of the $L$ target agents. 
In practice, we sample $\mathcal{J}_L$ based on the communication protocol $\bm{E}$, i.e., $\mathcal{J}_L\sim\frac{1}{\|\bm{E}_{i, :}\|_1}\bm{E}_{i,:}$, where $\bm{E}_{i,:}$ is the $i$-th row of $\bm{E}$.
Then, the agent $i$ computes a stochastic gradient $\sum_{j\in\mathcal{J}_L}\nabla_{\bm{u}^{(i)}}{f}_{\varepsilon}^{(i,j),t}$~\cite{zhao2014accelerated} and update $\bm{u}^{(i)}$ via
\begin{eqnarray}\label{eq:update_dual}
\begin{aligned}
    \bm{u}^{(i),t+1}\leftarrow \bm{u}^{(i),t} + \eta_t\sum_{j\in\mathcal{J}_L}\nabla_{\bm{u}^{(i)}}{f}_{\varepsilon}^{(i,j),t}.
\end{aligned}
\end{eqnarray}
Here, $\eta_t$ is the learning rate in the $t$-th iteration.
We set $\eta_t=\frac{\eta}{\sqrt{t+1}}$, where $\eta$ is the initial learning rate. 
The dual variables in the target agents can be updated in a similar way. 
Applying the above steps iteratively till the dual variables converge, each source agent $i$ can compute and store the local dual objectives $\{{f}_{\varepsilon}^{(i,j)}\}_{j=1}^{J}$ based on the information received during the iterations.
Accordingly, the source agent $i$ can compute the EOT distance $\widetilde{W}_{\varepsilon}(\mu,\gamma)$ by collecting $\{\{{f}_{\varepsilon}^{(i',j)}\}_{j=1}^{J}\}_{i'\neq i}$ from other source agents. 
Finally, the source agent $i$ broadcasts $\widetilde{W}_{\varepsilon}(\mu,\gamma)$ to all other agents. 

Algorithm~\ref{Algo-MRBCD} shows the MRBCD scheme, in which the communication complexity per step is given in red.
In particular, the kernel matrix $\bm{K}$ is represented as a set of sub-matrices, and the sub-matrices are stored in different agents: for each source agent $i$ (target agent $j$), we construct $\{\bm{K}_{ij}\}_{j=1}^{J}$ ($\{\bm{K}_{ij}\}_{i=1}^{I}$) by receiving data from the agents in the other domain though one-step communication, and the communication complexity of this step is $\mathcal{O}((IM+JN)D)$. 
When updating the dual variables, the communication cost per iteration is $\sum_{i\in I_L}N_i + \sum_{j\in J_L}M_j$.
When the numbers of samples in different agents are comparable, i.e., $N_i=\mathcal{O}(\frac{N}{I})$ and $M_j=\mathcal{O}(\frac{M}{J})$, the communication complexity per iteration can be represented as $\mathcal{O}(L(\frac{N}{I}+\frac{M}{J}))$. 
Accordingly, the overall communication complexity for updating dual variables is $\mathcal{O}(TL(\frac{N}{I}+\frac{M}{J}))$, where $T$ is the number of iterations. 
When $L=J$, we compute the gradient $\nabla_{\bm{u}^{(i)}}F_{\varepsilon}=\sum_{j=1}^{J}\nabla_{\bm{u}^{(i)}}{f}_{\varepsilon}^{(i,j)}$ exactly, and Algorithm~\ref{Algo-MRBCD} becomes the classic randomized block-coordinate descent (RBCD)~\cite{nesterov2012efficiency}. 
When $L=1$, we only consider the exchange of dual variables between an agent pair in each iteration. 
This setting is suitable for the agent with limited computation power because each iteration only involves a pair of agent.
Essentially, Algorithm~\ref{Algo-MRBCD} is a decentralized and mini-batch stochastic implementation of the randomized block-coordinate descent (RBCD) method~\cite{lu2015complexity,nesterov2012efficiency,richtarik2014iteration}. 

\subsection{Privacy-preserving Decentralization}

As shown in Algorithm~\ref{Algo-MRBCD}, the gradient of $\bm{u}^{(i)}$ (and that of $\bm{v}^{(j)}$) involves the construction of the kernel matrix $\bm{K}=\{\bm{K}_{ij}\}$, $j=1,...,J$ (and $i=1,...,I$), which requires us to transmit the raw data and the dual variables from one domain's agents to those in the other domain. 
The communication cost is high, especially for high-dimensional data.
Moreover, sharing raw data results in the leakage of private information, which is even infeasible in practical applications. 
Facing the above challenges, we consider the decentralized kernel approximation method in~\cite{khanduri2021decentralized}, constructing the kernel matrix without sharing raw data. 
Combining this method with our MRBCD scheme leads to the proposed DEOT method.

In particular, when the cost $c(\bm{x},\bm{y})$ is Euclidean, the kernel $\kappa(\bm{x},\bm{y})$ in~\eqref{eq:f-varepsilon} is a special case of the following generalized inner product (GIP) kernel~\cite{khanduri2021decentralized}:
\begin{eqnarray}\label{eq:GIP}
\begin{aligned}
    \kappa(\bm{x},\bm{y})=g(\phi(\bm{x},\bm{y}), \|\bm{x}\|,\|\bm{y}\|),
\end{aligned}
\end{eqnarray}
where $\phi(x,y)=\arccos(\frac{\langle \bm{x}, \bm{y}\rangle}{\|\bm{x}\|\|\bm{y}\|})$, and $g(\phi, \|\bm{x}\|,\|\bm{y}\|)$ is a $G$-Lipschitz continuous function with respect to $\phi$. 
According to the definition of the kernel function in~\eqref{eq:f-varepsilon}, we have
\begin{eqnarray}\label{eq:func_g}
    g(\phi, \|\bm{x}\|,\|\bm{y}\|)=\exp\Bigl(-\frac{1}{\varepsilon}(\|\bm{x}\|^2+\|\bm{y}\|^2-2\cos(\phi)\|\bm{x}\|\|\bm{y}\|)\Bigr).
\end{eqnarray}

\begin{algorithm}[t]
\caption{Privacy-preserving Kernel Approximation}\label{Algo-Kernel}
\begin{algorithmic}[1]
\STATE Draw random variables $\{\bm{\omega}_{\ell}\in\mathbb{R}^D\}_{\ell=1}^{Q}\sim \mathcal{N}(\bm{0}, \bm{I}_D)$ and broadcast them to all agents.
\FOR{Each source agent $i\in\{1,...,I\}$}
\STATE Construct $\bm{A}_{\mu_i}$ via~\eqref{eq:Amat} and broadcast it to all target agents.\hfill\textcolor{red}{$\mathcal{O}(JN_iQ)$}
\STATE If data is not normalized, broadcast $\{\|\bm{x}_{n}^{(i)}\|\}_{n=1}^{N_i}$ to all target agents.\hfill\textcolor{red}{$\mathcal{O}(JN_i)$}
\ENDFOR
\FOR{Each target agent $j\in\{1,...,J\}$}
\STATE Construct $\bm{A}_{\gamma_j}$ via~\eqref{eq:Amat} and broadcast it to all source agents.\hfill\textcolor{red}{$\mathcal{O}(IM_jQ)$}
\STATE If data is not normalized, broadcast $\{\|\bm{y}_{m}^{(j)}\|\}_{m=1}^{M_j}$ to all source agents.\hfill\textcolor{red}{$\mathcal{O}(IM_j)$}
\ENDFOR
\STATE Construct $\{\widehat{\bm{K}}_{ij}\}_{j=1}^{J}$ for each source agent $i$ and $\{\widehat{\bm{K}}_{ij}\}_{i=1}^{I}$ for each target agent $j$ via~\eqref{eq:Kmat}.
\end{algorithmic}
\end{algorithm}

For the GIP kernel, it is possible to approximate it without the share of raw data~\cite{khanduri2021decentralized}. 
Denote $D$ as the dimension of samples. 
Leveraging the random seed sharing method in~\cite{richards2020decentralised,xu2021coke}, we can sample $Q$ $D$-dimensional random variables from a multivariate normal distribution, i.e., $\{\bm{\omega}_{\ell}\}_{\ell=1}^{Q}\sim \mathcal{N}(\bm{0},\bm{I}_{D})$, and broadcast them to all the agents. 
Based on the random variables, we can construct a binary matrix for each agent.
Take the source agent $i$ as an example. 
Given $N_i$ samples $\{\bm{x}_{n}^{(i)}\}_{n=1}^{N_i}$, we have
\begin{eqnarray}\label{eq:Amat}
\begin{aligned}
    \bm{A}_{\mu_i}=[\mathbb{I}(\langle\bm{\omega}_{\ell}, \bm{x}_{n}^{(i)}\rangle\geq 0)]\in \{0,1\}^{Q\times N_i},
\end{aligned}
\end{eqnarray}
where $\mathbb{I}(\cdot)$ is an indicator, which outputs 1 if the input statement is true and outputs 0 otherwise. 
As a result, for each agent pair $(i,j)$, the kernel $\kappa(\bm{x}_{n}^{(i)},\bm{y}_{m}^{(j)})$ of their samples can be approximated by
\begin{eqnarray}\label{eq:Kmat}
\begin{aligned}
    \hat{\kappa}(\bm{x}_{n}^{(i)},\bm{y}_{m}^{(j)})
    &=g(\hat{\psi}(\bm{a}_{n}^{(i)},\bm{a}_{m}^{(j)}), \|\bm{x}_{n}^{(i)}\|, \|\bm{y}_{m}^{(j)}\|) \\
    &=g\Bigl(\pi \Bigl| 1 - \frac{2}{Q}\langle \bm{a}_{n}^{(i)},\bm{a}_{m}^{(j)}\rangle\Bigr|, \|\bm{x}_{n}^{(i)}\|, \|\bm{y}_{m}^{(j)}\|\Bigr),
\end{aligned}
\end{eqnarray}
where $\bm{a}_{n}^{(i)}$ is the $n$-th column of $\bm{A}_{\mu_i}$ and $\bm{a}_{m}^{(j)}$ is the $m$-th column of $\bm{A}_{\gamma_i}$.
Based on~\eqref{eq:Kmat}, we can obtain an approximated kernel matrix for an agent pair $(i, j)$, i.e., $\widehat{\bm{K}}_{ij}=[\hat{\kappa}(\bm{x}_{n}^{(i)},\bm{y}_{m}^{(j)})]$. 
This approximation preserves data privacy to some extent because it only requires two constructed binary matrices and the norms of samples. 

\begin{figure*}[t]
    \centering
    \includegraphics[width=1.0\linewidth]{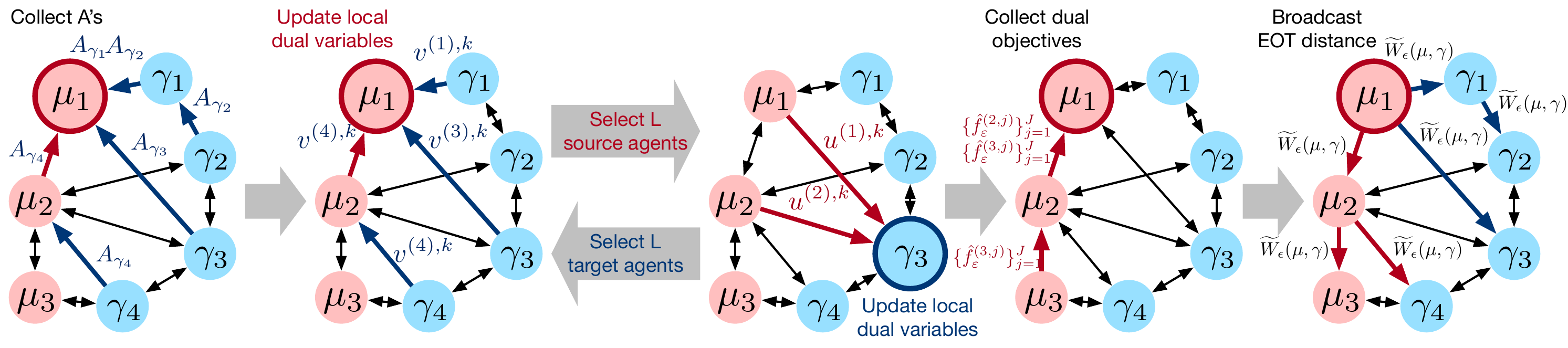}
    \caption{An illustration of the proposed DEOT method.}
    \label{fig:frame}
\end{figure*}

As shown in Algorithm~\ref{Algo-Kernel}, by one-shot communication, each agent obtains the matrices $\bm{A}$'s from all the agents in the other domain. 
Accordingly, the overall communication complexity is $\mathcal{O}((IM + JN)Q)$.
Note that, this complexity is independent with the sample dimension $D$, so it is suitable for high-dimensional cases. 
Moreover, even if $Q=\mathcal{O}(N)$, the practical communication cost can still be tractable because the matrices $\bm{A}$'s are binary and can be compressed before communication. 
Plugging the approximated kernel into~\eqref{eq:sample-eot}, we denote the dual objective using the approximated kernel, i.e., $F_{\varepsilon}(\bm{u},\bm{v};\widehat{\bm{K}},\bm{E})=\sum_{i,j}\frac{e_{ij}}{N_iM_j}f_{\varepsilon}(\bm{u}^{(i)},\bm{v}^{(j)};\widehat{\bm{K}}_{ij})$. 
As shown in Algorithm~\ref{Algo-DEOT}, the proposed DEOT method consists of two steps: $i)$ leveraging a theoretically-guaranteed method to approximate the kernel matrix without the share of raw data and $ii)$ updating the dual variables locally and iteratively in an MRBCD scheme. 
Fig.~\ref{fig:frame} further illustrates our DEOT method in details.

\begin{algorithm}[t]
\caption{Proposed DEOT Method}\label{Algo-DEOT}
\begin{algorithmic}[1] 
\STATE \textbf{Decentralized kernel approximation.}\hfill \textcolor{red}{$\mathcal{O}((IM+JN)Q)$}\\
For each source agent $i$ and target agent $j$, construct $\{\widehat{\bm{K}}_{ij}\}_{j=1}^{J}$ and $\{\widehat{\bm{K}}_{ij}\}_{i=1}^{I}$ via Algorithm~\ref{Algo-Kernel}.
\STATE \textbf{Decentralized entropic optimal transport.}\hfill \textcolor{red}{$\mathcal{O}(TL(\frac{N}{I}+\frac{M}{J})+IJ)$}\\
Update dual variables and compute $\max_{\bm{u},\bm{v}}F_{\varepsilon}(\bm{u},\bm{v};\widehat{\bm{K}},\bm{E})$ via Algorithm~\ref{Algo-MRBCD}.
\end{algorithmic}
\end{algorithm}

\subsection{Theoretical Analysis of Decentralized EOT}\label{ssec:theory}

The approximation error of the proposed DEOT method is determined by the following three factors: $i$) the mismatching between storage and communication protocols; $ii$) the perturbation on the optimization problem caused by the approximated kernel; and $iii)$ the sample complexity of the MRBCD scheme.
Taking the three factors into account, we can derive the bound of the expected approximation error under mild assumptions.
In particular, let $\mu=\sum_i p_i\mu_i$ and $\gamma=\sum_j q_j \gamma_j$ be the two distributions in a distributed system with $I$ source agents and $J$ target agents, whose storage and communication protocols are $\bm{pq}^{\top}=[p_iq_j]$ and $\bm{E}=[e_{ij}]$, respectively. 
We scatter $N$ samples of $\mu$ to the source agents and $M$ samples of $\gamma$ to the target agents. 
The kernel matrix of the samples is approximated as $\widehat{\bm{K}}$ via Algorithm~\ref{Algo-Kernel}, with the hyperparameter $P$.
Based on the communication protocol $\bm{E}$, let $\{\bm{u}^t,\bm{v}^t\}$ and $\{\hat{\bm{u}}^t,\hat{\bm{v}}^t\}$ denote the corresponding generated sequences based the exact kernel matrix $\bm{K}$ and $\widehat{\bm{K}}$ respectively, while $(\tilde{\bm{u}}^*,\tilde{\bm{v}}^*)$ and $(\hat{\bm{u}}^*,\hat{\bm{v}}^*)$ denote the optimal solutions based the exact kernel matrix $\bm{K}$ and the approximate one $\widehat{\bm{K}}$ respectively.
To summarize, the expectation of the approximation error can be upper-bound estimated by the corresponding errors concerning the above three factors, i.e.,
\begin{eqnarray}
\begin{aligned}
    {\cal{E}}_{total}&=\mathbb{E}_{\hat{\bm{u}}^t,\hat{\bm{v}}^t}\big[\big|F_{\varepsilon}(\hat{\bm{u}}^{t}, \hat{\bm{v}}^{t};\widehat{\bm{K}},\bm{E})-W_{\varepsilon}(\mu,\gamma) \big|\big]  \\
    &\leq \big| F_{\varepsilon}(\tilde{\bm{u}}^{*}, \tilde{\bm{v}}^{*};\bm{K},\bm{E}) - W_{\varepsilon}(\mu,\gamma) \big| + \big| F_{\varepsilon}(\hat{\bm{u}}^{*}, \hat{\bm{v}}^{*};\widehat{\bm{K}},\bm{E}) - F_{\varepsilon}(\tilde{\bm{u}}^{*}, \tilde{\bm{v}}^{*};\bm{K},\bm{E}) \big|  \\
    &\quad + \mathbb{E}_{\hat{\bm{u}}^t,\hat{\bm{v}}^t}\big[\big| F_{\varepsilon}(\hat{\bm{u}}^{t}, \hat{\bm{v}}^{t};\widehat{\bm{K}},\bm{E}) - F_{\varepsilon}(\hat{\bm{u}}^{*}, \hat{\bm{v}}^{*};\widehat{\bm{K}},\bm{E}) \big|\big]  \\
    &= \underbrace{\big| \widetilde{W}_{\varepsilon}(\mu,\gamma) - W_{\varepsilon}(\mu,\gamma) \big|}_{{\cal{E}}_{mismatching}} + \underbrace{\big| F_{\varepsilon}(\hat{\bm{u}}^{*}, \hat{\bm{v}}^{*};\widehat{\bm{K}},\bm{E}) - F_{\varepsilon}(\tilde{\bm{u}}^{*}, \tilde{\bm{v}}^{*};\bm{K},\bm{E}) \big|}_{{\cal{E}}_{kernel}}  \\
    & \quad + \underbrace{\mathbb{E}_{\hat{\bm{u}}^t,\hat{\bm{v}}^t}\big[\big| F_{\varepsilon}(\hat{\bm{u}}^{t}, \hat{\bm{v}}^{t};\widehat{\bm{K}},\bm{E}) - F_{\varepsilon}(\hat{\bm{u}}^{*}, \hat{\bm{v}}^{*};\widehat{\bm{K}},\bm{E}) \big|\big]}_{{\cal{E}}_{algorithm}}.  
\end{aligned}
\end{eqnarray}
${\cal{E}}_{mismatching}$, ${\cal{E}}_{kernel}$ and ${\cal{E}}_{algorithm}$ denote the mismatching error between storage and communication protocols, the perturbation error caused by the approximated kernel and the complexity error of the optimization algorithm, respectively.  
In the following, we will discuss the above three factors with respect to the approximation error respectively.

\medskip
\noindent - ${\cal{E}}_{mismatching}$: The mismatching error is denoted as the gap between $\widetilde{W}_{\varepsilon}(\mu, \gamma)$ and $W_{\varepsilon}(\mu, \gamma)$, which is irreducible when the storage and communication protocols are different.
We can establish the error through the following Lemma.
\begin{lemma}[Irreducible Estimation Error Caused by Mismatched Protocols.]\label{lm:protocol}
    Let $\mu=\sum_i p_i\mu_i$ and $\gamma=\sum_j q_j \gamma_j$ be the two distributions in a distributed system with $I$ source agents and $J$ target agents, whose storage and communication protocols are $\bm{pq}^{\top}=[p_iq_j]$ and $\bm{E}=[e_{ij}]$, respectively. 
    If $\max_{i,j}W_{\varepsilon}(\mu_i,\gamma_j)\leq \tau$ for some $\tau > 0$ and $\sum_{i,j}|e_{ij}-p_i q_j|\leq \sigma$ for some $\sigma >0$.
    We have
    \begin{eqnarray}
    |\widetilde{W}_{\varepsilon}(\mu, \gamma)-W_{\varepsilon}(\mu, \gamma)|\leq \tau\sigma.
    \end{eqnarray}
\end{lemma}
\begin{proof}
Let $u_1,v_1=\arg\sup_{u, v \in \mathcal{C}(\mathcal{X})} \mathbb{E}_{(i, j)\sim \bm{E}}[\mathbb{E}_{\bm{x}\sim \mu_i,\bm{y}\sim\gamma_j}[f_{\varepsilon}(u,v;\kappa(\bm{x},\bm{y}))]]$ be the optimal dual functions of $W_{\varepsilon}(\mu, \gamma)$.
Similarly, let $u_2,v_2$ be the optimal dual functions of $\widetilde{W}_{\varepsilon}(\mu, \gamma)$.
We have
\begin{eqnarray*}
\begin{aligned}
    &|\widetilde{W}_{\varepsilon}(\mu, \gamma)-W_{\varepsilon}(\mu, \gamma)| \\
    \leq &
    \begin{cases}
        \sum_{i,j}(e_{ij}-p_i q_j)\mathbb{E}_{\bm{x}\sim \mu_i,\bm{y}\sim\gamma_j}[f_{\varepsilon}(u_1,v_1;\kappa(\bm{x},\bm{y}))] &\text{if}~\widetilde{W}_{\varepsilon}(\mu, \gamma)\geq W_{\varepsilon}(\mu, \gamma)\\
        \sum_{i,j}(p_i q_j - e_{ij})\mathbb{E}_{\bm{x}\sim \mu_i,\bm{y}\sim\gamma_j}[f_{\varepsilon}(u_2,v_2;\kappa(\bm{x},\bm{y}))] &\text{if}~\widetilde{W}_{\varepsilon}(\mu, \gamma)< W_{\varepsilon}(\mu, \gamma)\\
    \end{cases}\\
    \leq & 
    \sideset{}{_{i,j}}\sum|e_{ij}-p_i q_j|\sideset{}{_{u\in\{u_1,u_2\},v\in\{v_1,v_2\}}}\max\mathbb{E}_{\bm{x}\sim \mu_i,\bm{y}\sim\gamma_j}[f_{\varepsilon}(u,v;\kappa(\bm{x},\bm{y}))]\\
    \leq &
    \sideset{}{_{i,j}}\sum|e_{ij}-p_i q_j|\sideset{}{_{u,v\in\mathcal{C}_{\mathcal{X}}}}\sup\mathbb{E}_{\bm{x}\sim \mu_i,\bm{y}\sim\gamma_j}[f_{\varepsilon}(u,v;\kappa(\bm{x},\bm{y}))]\\
    = &
    \sideset{}{_{i,j}}\sum|e_{ij}-p_i q_j|W_{\varepsilon}(\mu_i, \gamma_j)\\
    \leq & \sideset{}{_{i,j}}\max W_{\varepsilon}(\mu_i, \gamma_j)\sideset{}{_{i,j}}\sum|e_{ij}-p_i q_j|\\
    \leq & \tau\sigma, 
\end{aligned}
\end{eqnarray*}which indicates the result of this Lemma.
\end{proof}
{\sc{Lemma}}~\ref{lm:protocol} indicates that as long as the mismatch between the storage and communication protocols (i.e., $\sigma$) is small, we can approximate $W_{\varepsilon}(\mu, \gamma)$ well by $\widetilde{W}_{\varepsilon}(\mu, \gamma)$.\footnote{{\sc{Lemma}}~\ref{lm:protocol} is valid for both continuous probability measures and sample-based discrete measures.}

\medskip
\noindent - ${\cal{E}}_{kernel}$: The perturbation error caused by the approximated kernel is considered as the optimal function value error with respect to the kernel matrix perturbation.
The approximated kernel matrix is calculated through Algorithm~\ref{Algo-Kernel}, so that we firstly need to estimate the distance between the exact kernel matrix $\bm{K}$ and the obtained approximated kernel matrix $\widehat{\bm{K}}$.
According to~\cite{khanduri2021decentralized}, we can directly obtain the following theoretical result.
\begin{lemma}[Approximation Error of Kernel~\cite{khanduri2021decentralized}]\label{thm:kernel-approx}
Let $\bm{K}\in\mathbb{R}^{N\times M}$ be the matrix defined by the GIP kernel in~\eqref{eq:GIP} and $\widehat{\bm{K}}$ be the approximation achieved via~\eqref{eq:Kmat}, with probability at least $1-\delta$, we have
\begin{eqnarray}
    \|\bm{K}-\widehat{\bm{K}}\|\leq G(N+M)\Bigl( \sqrt{\frac{32\pi^2}{Q}\log\frac{2(N+M)}{\delta}} + \frac{8\pi}{3Q}\log\frac{2(N+M)}{\delta} \Bigr).
\end{eqnarray}
\end{lemma}
\begin{proof}
The $N\times M$ kernel matrix $\bm{K}$ can be considered as a sub-block matrix of the $(N+M)\times (N+M)$ full kernel matrix based on given $N+M$ data samples.
This full kernel matrix can be denoted as $\bm{K}_{full}\in {\mathbb{R}}^{(N+M)\times (N+M)}$.
Accordingly, the proposed Algorithm~\ref{Algo-Kernel} can be considered as a partial version of the Algorithm 1 in~\cite{khanduri2021decentralized}.
The approximate level of $\widehat{\bm{K}}_{full} \in {\mathbb{R}}^{(N+M)\times (N+M)}$ obtained through the Algorithm 1 in~\cite{khanduri2021decentralized} has been proven in Lemma 4.1 in~\cite{khanduri2021decentralized}: With probability at least $1-\delta$, we have
\begin{eqnarray*}\label{eq:kernel-error2}
\begin{aligned}
    \|\bm{K}_{full} - \widehat{\bm{K}}_{full}\|
    \leq G(N+M)\Bigl( \sqrt{\frac{32\pi^2}{Q}\log\frac{2(N+M)}{\delta}} + \frac{8\pi}{3Q}\log\frac{2(N+M)}{\delta} \Bigr).
\end{aligned}
\end{eqnarray*}
Because $\|\bm{K} - \widehat{\bm{K}}\| \leq \|\bm{K}_{full} - \widehat{\bm{K}}_{full}\|$, we derive the result of this Lemma.
\end{proof}

Each $f_{\varepsilon}(u,v;\kappa(\bm{x},\bm{y}))$ is typically convex and Lipschitz continuous with respect to $(u, v)$~\cite{genevay2016stochastic}.
Therefore, the objective function $F_{\varepsilon}(\bm{u},\bm{v};\bm{K},\bm{E})$ is Lipschitz continuous with respect to $(\bm{u},\bm{v})$.
Further based on the definition of $f_{\varepsilon}^{(i,j)}$, it is obvious that $f_{\varepsilon}^{(i,j)}$ is a liner function with respect to $\kappa(\bm{x}_n^{(i)},\bm{y}_m^{(j)})$.
On the whole, the objective function $F_{\varepsilon}$ can be considered as a linear function with respect to kernel matrix $\bm{K}$ and thus also is Lipschitz continuous with respect to $\bm{K}$.
According to Lemma 3.1 in~\cite{dempe2015lipschitz}, if we model $\bm{K}$ as the variable of the parametric optimization problem
\begin{eqnarray}
    \phi(\bm{K}) = \max_{\bm{u},\bm{v}} F_{\varepsilon} \big(\bm{u}, \bm{v}; \bm{K}, \bm{E} \big).
\end{eqnarray}
We can conclude that the optimal value function $\phi\big(\bm{K}\big)$ with respect to $\bm{K}$ is $L_{\kappa}$-Lipschitz continuous, i.e.,
\begin{eqnarray}\label{eq:error_kernel_approx}
\begin{aligned}
    &\big| F_{\varepsilon}(\hat{\bm{u}}^{*}, \hat{\bm{v}}^{*};\widehat{\bm{K}},\bm{E}) - F_{\varepsilon}(\tilde{\bm{u}}^{*}, \tilde{\bm{v}}^{*};\bm{K},\bm{E})  \big| \\ 
=& \big| \phi(\widehat{\bm{K}}) - \phi(\bm{K}) \big| \\
\leq& L_{\kappa} \big\| \widehat{\bm{K}} - \bm{K} \big\| \\
\leq& L_{\kappa} G(N+M)\Bigl( \sqrt{\frac{32\pi^2}{Q}\log\frac{2(N+M)}{\delta}} + \frac{8\pi}{3Q}\log\frac{2(N+M)}{\delta} \Bigr).
\end{aligned}
\end{eqnarray}
The above inequality can be considered as the upper bound approximation of ${\cal{E}}_{kernel}$.

\medskip
\noindent - ${\cal{E}}_{algorithm}$: The complexity error of the proposed algorithm is considered as the iteration complexity of Algorithm~\ref{Algo-MRBCD}.
MRBCD is a typical stochastic first order method, which has been theoretically discussed in~\cite{Dang2015,Hu2022} for general cases.
The employed MRBCD can be considered as the Algorithm 2 in~\cite{Hu2022}, while the variable block is chosen randomly in each iteration and samples are chosen following the mini-batch scheme.
Following the Theorem 5 and Corollary 4 in~\cite{Hu2022}, we can obtain the iteration complexity result of our proposed MRBCD Algorithm~\ref{Algo-MRBCD} in the following Lemma.
\begin{lemma}\label{le:algorithm-error}
For the problem in~\eqref{eq:sample-eot} with a kernel matrix $\bm{K}$, let $(\bm{u}^*,\bm{v}^*)\in \mathcal{C}^*$ be the optimal solution in the optimal solution set and $\left\{ \left(\bm{u}^t,\bm{v}^t\right) \right\}$ be the sequence generated by Algorithm~\ref{Algo-MRBCD}.
With properly chosen parameter $\eta$, we have
\begin{eqnarray}\label{eq:convergence-theorem}
\begin{aligned}
    \qquad \mathbb{E} \big[ |F_{\varepsilon}(\hat{\bm{u}}^{t},\hat{\bm{v}}^{t};\bm{K},\bm{E})-F_{\varepsilon}(\bm{u}^*,\bm{v}^*;\bm{K},\bm{E})| \big]
    \leq \mathcal{O}\Bigl(\frac{1}{\sqrt{t}}\Bigr),
\end{aligned}
\end{eqnarray}
where the expectation is calculated with respect to the randomly-selected agents.
\end{lemma}
The above {\sc{Lemma}}~\ref{le:algorithm-error} shows the iteration complexity of the proposed algorithm and indicates the upper bound of the complexity error ${\cal{E}}_{algorithm}$.

%% file: sections-arxiv/extension.tex
\section{An Extension to Decentralized Entropic Gromov-Wasserstein}

\subsection{Dual Formulation of EGW Distance}

\smallskip
When the data of the two domains are in two incomparable metric-measure spaces, we need to compute the entropic Gromov-Wasserstein (EGW) distance between them~\cite{peyre2016gromov,zhang2022gromov} in a decentralized way, achieving privacy-preserving and communication efficiency jointly. 
Fortunately, our method can be extended to achieve this aim. 
In particular, suppose that we have two metric measure spaces, denoted as $(\mathcal{X},d_{\mathcal{X}},\mu)$ and $(\mathcal{Y},d_{\mathcal{Y}},\gamma)$. 
Due to the shift-invariance of EGW distance, we can assume $\int_{\mathcal{X}}\bm{x}\mu(\bm{x})\mathrm{d}\bm{x}=\bm{0}$ and $\int_{\mathcal{Y}}\bm{y}\gamma(\bm{y})\mathrm{d}\bm{y}=\bm{0}$, respectively, without the loss of generality. 
The (squared) entropic Gromov-Wasserstein distance can be defined as follows:
\begin{eqnarray}\label{eq:egw}
\begin{aligned}
    &GW_{\varepsilon}(\mu,\gamma) \stackrel{\text {def.}}{=}\\ 
    &\inf_{\pi \in \Pi(\mu, \gamma)} \int_{\mathcal{X}^2\times\mathcal{Y}^2} |d_{\mathcal{X}}^2(\bm{x},\bm{x}')-d_{\mathcal{Y}}^2(\bm{y},\bm{y}')|^2\pi(\bm{x}, \bm{y})\pi(\bm{x}', \bm{y}') \mathrm{d}\bm{x}\mathrm{d}\bm{x}'\mathrm{d}\bm{y}\mathrm{d}\bm{y}' - \varepsilon H(\pi). 
\end{aligned}
\end{eqnarray}
Based on the work in~\cite{zhang2022gromov}, we can derive the dual form of entropic Gromov-Wasserstein (EGW) distance when $\mathcal{X}\subset\mathbb{R}^{D_X}$, $\mathcal{Y}\subset\mathbb{R}^{D_Y}$, and $d_{\mathcal{X}}$ and $d_{\mathcal{Y}}$ are Euclidean distance:
\begin{eqnarray}\label{eq:dual_egw}
\begin{aligned}
    &GW_{\varepsilon}(\mu,\gamma) \\ =&\underbrace{\int_{\mathcal{X}^2}d_{\mathcal{X}}^4(\bm{x},\bm{x}')\mu(\bm{x})\mu(\bm{x}')\mathrm{d}\bm{x}\mathrm{d}\bm{x}'}_{S_X}+\underbrace{\int_{\mathcal{Y}^2}d_{\mathcal{Y}}^4(\bm{y},\bm{y}')\gamma(\bm{y})\gamma(\bm{y}')\mathrm{d}\bm{y}\mathrm{d}\bm{y}'}_{S_Y} \\
&-\underbrace{4\int_{\mathcal{X}\times\mathcal{Y}}\|\bm{x}\|_2^2\|\bm{y}\|_2^2\mu(\bm{x})\gamma(\bm{y})\mathrm{d}\bm{x}\mathrm{d}\bm{y}}_{S_{XY}}+\sideset{}{}\inf_{\bm{P}\in\mathbb{R}^{D_X\times D_Y}}\sideset{}{}\sup_{u\in\mathcal{C}_{\mathcal{X}},v\in\mathcal{C}_{\mathcal{Y}}} \Bigl(32\|\bm{P}\|_F^2+  \\
&\int_{\mathcal{X}} u(\bm{x})\mu(\bm{x}) \mathrm{d}\bm{x}  + \int_{\mathcal{Y}} v(\bm{y})\gamma(\bm{y}) \mathrm{d}\bm{y} -\varepsilon \int_{\mathcal{X}\times\mathcal{Y}} e^{\frac{u(\bm{x})+v(\bm{y})}{\varepsilon}}\underbrace{e^{\frac{-c(\bm{x}, \bm{y};\bm{P})}{\varepsilon}}}_{\kappa(\bm{x}, \bm{y};\bm{P})}\mu(\bm{x})\gamma(\bm{y}) \mathrm{d}\bm{x}\mathrm{d}\bm{y} \Bigr)  \\
=&S_x+S_y-S_{xy}+\sideset{}{}\inf_{\bm{P}\in\mathbb{R}^{D_X\times D_Y}}\sideset{}{}\sup_{u\in\mathcal{C}_{\mathcal{X}},v\in\mathcal{C}_{\mathcal{Y}}} 32\|\bm{P}\|_F^2+\mathbb{E}_{\bm{x}\sim\mu,\bm{y}\sim\gamma}f_{\varepsilon}(u,v;\kappa(\bm{x},\bm{y};\bm{P})), 
\end{aligned}
\end{eqnarray}
where $\bm{P}\in\mathbb{R}^{D_X\times D_Y}$ is an bilinear alignment matrix, and $f_{\varepsilon}(u,v;\kappa(\bm{x},\bm{y};\bm{P}))$ is defined the same with that in~\eqref{eq:f-varepsilon}, in which the cost $c$ associated with the kernel $\kappa$ is parametrized by $\bm{P}$ as
\begin{eqnarray}\label{eq:cost_egw}
\begin{aligned}
    c(\bm{x},\bm{y};\bm{P}) = -4\|\bm{x}\|_2^2\|\bm{y}\|_2^2-32\bm{x}^{\top}\bm{Py}.
\end{aligned}
\end{eqnarray}

Similar to~\eqref{eq:dual-eot2}, in a decentralized scenario, in which the samples of $\mu$ and $\gamma$ are scattered to different agents and the agents communicate with each other under the protocol $\bm{E}$, we can reformulate the optimization problem in~\eqref{eq:dual_egw} as
\begin{eqnarray}\label{eq:dual_egw2}
    \sideset{}{}\inf_{\bm{P}\in\mathbb{R}^{D_X\times D_Y}}\sideset{}{}\sup_{u\in\mathcal{C}_{\mathcal{X}},v\in\mathcal{C}_{\mathcal{Y}}} 32\|\bm{P}\|_F^2+\mathbb{E}_{(i,j)\sim \bm{E}}\mathbb{E}_{\bm{x}\sim\mu_i,\bm{y}\sim\gamma_j}f_{\varepsilon}(u,v;\kappa(\bm{x},\bm{y};\bm{P})),
\end{eqnarray}
leading to the proposed decentralized entropic Gromov-Wasserstein (DEGW) problem.

Given the samples of $\mu$ and $\gamma$, we can further reformulate~\eqref{eq:dual_egw2} as the following min-max optimization problem:
\begin{eqnarray}\label{eq:sample_egw}
\begin{aligned}
    \min_{\bm{P}\in\mathbb{R}^{D_X\times D_Y}}\max_{\substack{\bm{u}=\{\bm{u}^{(i)}\}_{i=1}^{I} \in \mathbb{R}^{N}\\ \bm{v}=\{\bm{v}^{(j)}\}_{j=1}^{J} \in \mathbb{R}^{M}}}
    32\|\bm{P}\|_F^2+F_{\varepsilon}(\bm{u},\bm{v};\bm{K}(\bm{P}),\bm{E}),
\end{aligned}
\end{eqnarray}
where $F_{\varepsilon}$ is defined as the objective function in~\eqref{eq:sample-eot}, and kernel matrix is parametrized by $\bm{P}$, i.e., $\bm{K}(\bm{P})=[\exp(-\frac{c(\bm{x},\bm{y};\bm{P})}{\varepsilon})]$.

Ignoring privacy protection, we can solve the DEGW problem in~\eqref{eq:sample_egw} by alternating optimization.
Specifically, in each step, we first fix $\bm{P}$ and update the dual variables by block coordinate descent scheme. 
Then, we can fix the dual variable and update $\bm{P}$ by gradient descent, in which
\begin{eqnarray}\label{eq:grad_A}
\begin{aligned}
    &\nabla_{\bm{P}}F_{\varepsilon}(\bm{u},\bm{v};\bm{K}(\bm{P}),\bm{E})\\
    =&-\sum_{i=1}^{I}\sum_{j=1}^{J}\frac{e_{ij}}{N_iM_j}\underbrace{\sum_{n=1}^{N_i}\sum_{m=1}^{M_j}\varepsilon e^{\frac{\bm{u}_{n}^{(i)} + \bm{v}_m^{(j)}}{\varepsilon}}\kappa(\bm{x}_n^{(i)}, \bm{y}_m^{(j)};\bm{P})\bm{x}_n^{(i)}(\bm{y}_m^{(j)})^{\top}}_{\nabla_{\bm{P}}f_{\varepsilon}^{(i,j)}}.
\end{aligned}
\end{eqnarray}
Furthermore, we can update $\bm{P}$ by mini-batch randomization scheme as well, i.e., computing the gradient randomly based on the data in a specific agent and broadcasting the updated $\bm{P}$ to other agents. 
Accordingly, given the optimal $\bm{u}^*$, $\bm{v}^*$, and $\bm{P}^*$, we can obtain $\widetilde{GW}_{\varepsilon}(\mu,\gamma)=S_X+S_Y-S_{XY}+F_{\varepsilon}(\bm{u}^*,\bm{v}^*;\bm{K}(\bm{P}^*),\bm{E})$. 

To emphasize, in this case, we can not directly employ the theoretical result in {\sc{Lemma}}~\ref{le:algorithm-error}, but the convergence and iteration complexity can also be obtained.
Our algorithm can be treated as a special case of the proposed BAPG algorithm in~\cite{xu2023unified}, while only the dual variables are updated through the block coordinate scheme.
Following~\cite[Theorem 5.3 and Theorem 5.4]{xu2023unified}, we could obtain our algorithm's convergence and iteration complexity results.
Besides the convergence result, the iteration complexity to obtain an $\epsilon$-stationary point for problem \eqref{eq:sample_egw} can be bounded by ${\cal{O}}(\epsilon^{-4})$.

\subsection{Privacy-preserving Decentralization}

When sharing raw data is forbidden, solving the DEGW problem becomes challenging because $i)$ we need to approximate $S_X$, $S_Y$, $S_{XY}$, and the kernel $\bm{K}(\bm{P})$ while their computations require raw data, and more importantly, $ii)$ we need to compute the gradient of $\bm{P}$ in a way differing from~\eqref{eq:grad_A}. 
To achieve this aim, we modify the above kernel approximation method so that it can estimate the components in the DEGW problem.
In particular, suppose that each agent $i$ in the source domain receives the binary data $\bm{A}_{\gamma_j}=[\bm{a}_{m}^{(j)}]$ and the norm of raw data $\{\|\bm{y}_{m}^{(j)}\|\}_{m=1}^{M_j}$ from the agent $j$ in the target domain, and the binary data and norms from the agent $i'$ in the target domain. 

\medskip
\noindent - {\bf{Computation of $S_{XY}$}}. 
As shown in~\eqref{eq:dual_egw}, the computation of $S_{XY}$ only involves the norm of raw data, which can be achieved directly based on the received data norm.
Specifically, when the agent $i$ receives $\{\|\bm{y}_{m}^{(j)}\|\}_{m=1}^{M_j}$, it can compute the sub-matrix $S_{XY}^{(i,j)}\in\mathbb{R}^{N_i\times M_j}$, whose element is
\begin{eqnarray}\label{eq:sxy}
    S_{XY}^{(i,j)}(n,m)=\frac{4e_{ij}}{N_i M_j}\|\bm{x}_{n}^{(i)}\|\|\bm{y}_{m}^{(j)}\|,~\forall n=1,...,N_i,~m=1,...,M_j.
\end{eqnarray}

\medskip
\noindent - {\bf{Approximation of $S_X$ and $S_Y$}}. 
Recall the kernel approximation in~\eqref{eq:Kmat}.
The term $\pi | 1 - \frac{2}{Q}\langle \bm{a}_{n}^{(i)},\bm{a}_{m}^{(j)}\rangle|$ actually works for approximating $\arccos \frac{\langle \bm{x}_{n}^{(i)}, \bm{x}_{m}^{(j)}\rangle}{\|\bm{x}_{n}^{(i)}\|_2\|\bm{x}_{m}^{(j)}\|_2}$. 
Therefore, based on received binary vectors $\bm{A}_{\mu_{i'}}=[\bm{a}_{n}^{(i')}]$ and norms $\{\|\bm{x}_{n}^{(i')}\|\}_{n=1}^{N_{i'}}$, we can approximate the Euclidean distance between original $\bm{x}$'s, and accordingly, the sub-matrix of $S_{X}$, i.e., $S_{X}^{(i,i')}$ can be approximated as
\begin{eqnarray}\label{eq:sx}
\begin{aligned}
    &S_{X}^{(i,i')}(n,n') \\
    = &\frac{e_ie_{i'}}{N_i N_{i'}}\Bigl(\underbrace{\|\bm{x}_{n}^{(i)}\|_2^2+\|\bm{x}_{n'}^{(i')}\|_2^2-2\cos\Bigl(\pi \Bigl| 1 - \frac{2}{Q}\langle \bm{a}_{n}^{(i)},\bm{a}_{n'}^{(i')}\rangle\Bigr|\Bigr)\|\bm{x}_{n}^{(i)}\|_2\|\bm{x}_{n'}^{(i')}\|_2}_{d_{\mathcal{X}}^2(\bm{x}_n^{(i)},\bm{x}_{n'}^{(i')})}\Bigr)^2, \\
    &\forall n=1,\cdots,N_i,\quad \forall n'=1,\cdots,N_{i'},
\end{aligned}
\end{eqnarray}
where $e_i=\sum_{j=1}^J e_{ij}$ is the probability of selecting the source agent $i$ based on the communication protocol.
For $S_Y$, its sub-matrices $\{S_{Y}^{(j,j')}\}_{j,j'=1}^{J}$ can be approximated in the same way.

\medskip
\noindent - {\bf{Approximate $\bm{K}(\bm{A})$}}. 
The kernel function in EGW distance is $$\exp\Bigl(\frac{4\|\bm{x}\|_2^2\|\bm{y}\|_2^2+32\bm{x}^{\top}\bm{Py}}{\varepsilon}\Bigr),$$
which involves the inner product $\langle \bm{P}^{\top}\bm{x}, \bm{y}\rangle$. 
Therefore, we can construct the binary vector for each $\bm{P}^{\top}\bm{x}$, i.e., $\bm{b}_{n}^{(i)}=[\mathbb{I}(\langle \bm{\omega}_{\ell}, \bm{P}^{\top}\bm{x}_{n}^{(i)}\rangle \geq 0)]\in\{0, 1\}^{Q}$ for $n=1,...,N_i$, and estimate the kernel as
\begin{eqnarray}\label{eq:Kmat2}
\begin{aligned}
    &\hat{\kappa}(\bm{x}_{n}^{(i)},\bm{y}_{m}^{(j)};\bm{P})\\
    =&\exp\Bigl(\frac{4\|\bm{x}_{n}^{(i)}\|_2^2\|\bm{y}_{m}^{(j)}\|_2^2}{\varepsilon}+\frac{32}{\varepsilon}\cos\Bigl(\pi \Bigl| 1 - \frac{2}{Q}\langle \bm{b}_{n}^{(i)},\bm{a}_{m}^{(j)}\rangle\Bigr|\Bigr)\|\bm{P}^{\top}\bm{x}_{n}^{(i)}\|_2\|\bm{y}_{m}^{(j)}\|_2\Bigr),\\
        & \forall n=1,\cdots,N_i,\,\, \forall m=1,\cdots,M_j. 
\end{aligned}
\end{eqnarray}
For the agent $i$, we can construct $\widehat{\bm{K}}_{ij}(\bm{P})=[\hat{\kappa}(\bm{x}_{n}^{(i)},\bm{y}_{m}^{(j)};\bm{P})]$ based on received binary matrix $\bm{A}_{\gamma_j}=[\bm{a}_m^{(j)}]$ and norms $\{\|\bm{y}_m^{(j)}\|\}_{m=1}^{M_j}$. 
The collection of all sub-matrices $\{\widehat{\bm{K}}_{ij}(\bm{P})\}$ leads to the approximated kernel $\widehat{\bm{K}}(\bm{P})$.

\medskip
\noindent - {\bf{Approximate the gradient $\nabla_{\bm{P}}\widehat{\bm{K}}(\bm{P})$}}.  
When $\bm{P}^{\top}\bm{x}_n^{(i)}$ is replaced by the binary vector $\bm{b}_n^{(i)}$, the gradient $\nabla_{\bm{P}}\widehat{\bm{K}}(\bm{P})$ becomes infeasible because $\bm{b}_n^{(i)}$ is non-differentiable.
To solve this issue, we replace the binary vector $\bm{b}_n^{(i)}$ in~\eqref{eq:Kmat2} with the Sigmoid function, i.e., $$\sigma_{n}^{(i)}=[{1}/({1+\exp(-\langle\bm{\omega}_{\ell},\bm{P}^{\top}\bm{x}_{n}^{(i)}\rangle)})]\in [0,1]^{Q},$$when computing the gradient.
As a result, we can approximate the gradient $\nabla_{\bm{P}}\widehat{\bm{K}}(\bm{P})$ based on the chain rule of the gradient of the composite function.

In summary, the scheme of our DEGW method is shown in Algorithm~\ref{Algo-MRBCD2}.
Here, an alternating optimization strategy is applied to update the dual variables and the bilinear alignment matrix, and the number of alternating optimization steps is indicated by $T_{outer}$.
When updating the dual variables, we apply Algorithm~\ref{Algo-MRBCD} with $T_{inner}$ iterative steps.
In Algorithm~\ref{Algo-MRBCD2}, the communication complexity of each step is shown in red. 

\begin{algorithm}[t]
\caption{Proposed DEGW Method}\label{Algo-MRBCD2}
\begin{algorithmic}[1] 
\STATE For each source agent $i$ and target agent $j$, receive binary vectors and norms from the agents in the other domain, and initialize $\bm{u}^{(i),0}=\bm{0}$ and $\bm{v}^{(j),0}=\bm{0}$.\hfill \textcolor{red}{$\mathcal{O}(IMQ+JNQ)$}
\STATE For a specific agent, initialize $\bm{P}^0$ randomly.
\FOR{$t=0,1,\cdots,T_{outer}$}
\STATE Set the learning rate $\eta_t=\frac{\eta}{\sqrt{t+1}}$.
\STATE \textbf{1) Update $\bm{P}$ and broadcast it:}
\FOR{An agent pair $(i,j)\sim \bm{E}$}
    \STATE Agent $i$ constructs $\widehat{\bm{K}}_{ij}(\bm{P}^t)$ and approximates $\nabla_{\bm{P}}\widehat{\bm{K}}_{ij}(\bm{P}^t)$ accordingly.
    \STATE $\nabla_{\bm{P}}f_{\varepsilon}^{(i,j),t}=\nabla_{\bm{P}}\widehat{\bm{K}}_{ij}(\bm{P}^t)\nabla_{\widehat{\bm{K}}_{ij}}f_{\varepsilon}^{(i,j),t}$, and $\bm{P}^{t+1}\leftarrow \bm{P}^{t}-\eta_t\sum_{i\in\mathcal{I}_L}\nabla_{\bm{P}}f_{\varepsilon}^{(i,j),t}$
    \STATE Agent $i$ broadcasts $\bm{P}^{(t+1)}$ to the other agents.\hfill \textcolor{red}{$\mathcal{O}((I+J)D_XD_Y)$} 
\ENDFOR
\STATE \textbf{2) Update dual variables via our MRBCD scheme:}\hfill \textcolor{red}{$\mathcal{O}(T_{inner}L(\frac{N}{I}+\frac{M}{J}))$}\\
$\{\bm{u}^{(i),t+1},\bm{v}^{(j),t+1}\}_{i,j=1}^{I,J}=\arg\max_{\bm{u},\bm{v}}F_{\varepsilon}(\bm{u},\bm{v};\widehat{\bm{K}}(\bm{P}^t))$ via Algorithm~\ref{Algo-MRBCD}.
\ENDFOR
\STATE For an arbitrary source agent $i$, receive binary vectors and norms from the other agents in the same domain and compute $S_X$ via~\eqref{eq:sx} accordingly.\hfill \textcolor{red}{$\mathcal{O}(NQ)$}
\STATE For an arbitrary target agent $j$, receive binary vectors and norms from the other agents in the same domain and compute $S_Y$ via~\eqref{eq:sx} accordingly.\hfill \textcolor{red}{$\mathcal{O}(MQ)$}
\STATE Based on the received norms, the source agent $i$ computes $S_{XY}$ via~\eqref{eq:sxy}.
\STATE The source agent $i$ receives the optimal dual objectives $\{\{\hat{f}_{\varepsilon}^{(i',j)}\}_{j=1}^{J}\}_{i'\neq i}$ from the remaining agents in the source domain.\hfill \textcolor{red}{$\mathcal{O}(IJ)$}
\STATE Compute $\widetilde{GW}_{\varepsilon}(\mu,\gamma)=S_X+S_Y-S_{XY}+\sum_{i,j}\hat{f}_{\varepsilon}$ in the source agent $i$ and broadcast it to all other agents.\hfill \textcolor{red}{$\mathcal{O}(IJ)$}
\end{algorithmic}
\end{algorithm}


%% file: sections-arxiv/exp.tex
\section{Numerical Experiments}\label{sec:exp}
To demonstrate the effectiveness of our decentralized EOT method, we analyze its performance on synthetic data and apply it to distributed domain adaptation tasks.

\subsection{Analytic Experiments on Synthetic Data}

We consider two synthetic datasets in this experiment: the first dataset contains two 5-dimensional Gaussian distributions $(\mathcal{N}_1,\mathcal{N}_2)$, each of which includes 2,000 samples, and the second one contains two 5-dimensional Gaussian mixture models $(\mathcal{M}_1,\mathcal{M}_2)$, each of which includes two Gaussian components and 2,000 samples.
For each dataset, we randomly scatter one distribution's samples to eight source agents and the other distribution's samples to eight target agents, respectively. 
When scattering the samples of the Gaussian mixture models (GMMs), we apply two strategies: $i)$ scattering the samples randomly to the agents such that different agents store \textbf{i.i.d.} samples, and $ii)$ each agent stores the samples of a single Gaussian component such that different agents store \textbf{non-i.i.d.} samples. 
Following the decentralized optimization work in~\cite{hughes2021fair,zhang2019consensus}, we assume the network of the agents to be connected, i.e., a route always exists between two arbitrary agents. 
Accordingly, the storage and communication protocols are uniform distributions.

\begin{figure}[t!]
    \centering
    \subfigure[$W_{\varepsilon}(\mathcal{N}_1,\mathcal{N}_2)$, $L=1$]{
    \includegraphics[height=4.0cm]{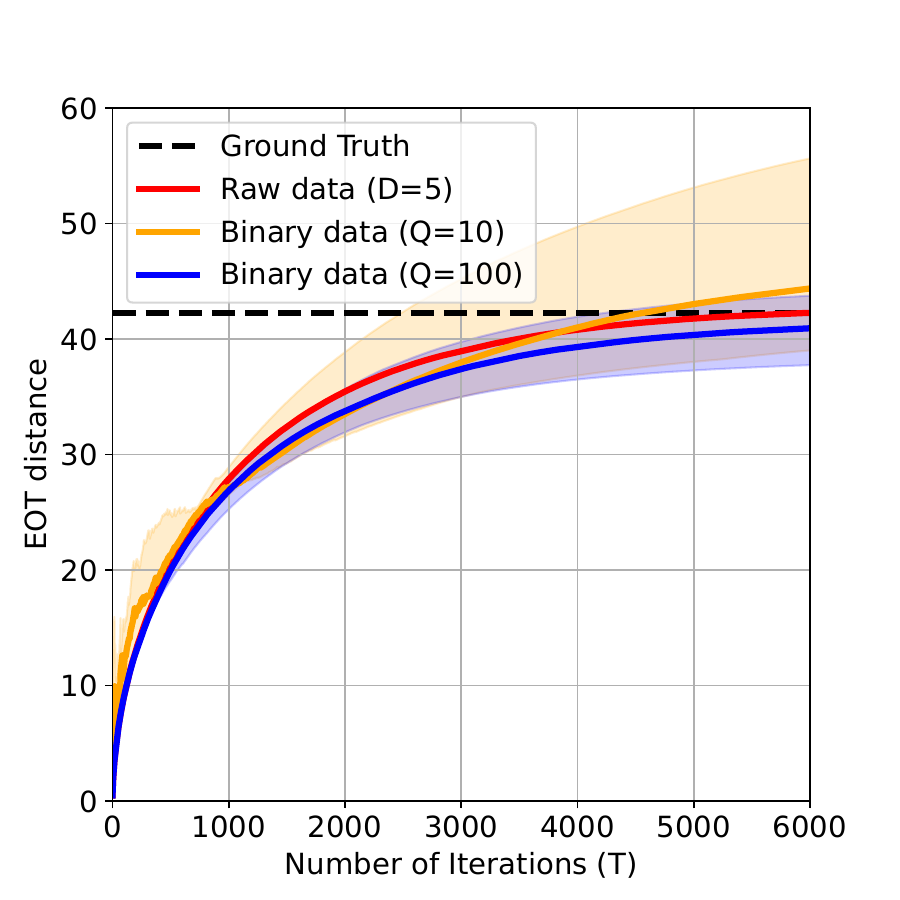}
    }
    \subfigure[$W_{\varepsilon}(\mathcal{N}_1,\mathcal{N}_2)$, $L=4$]{
    \includegraphics[height=4.0cm]{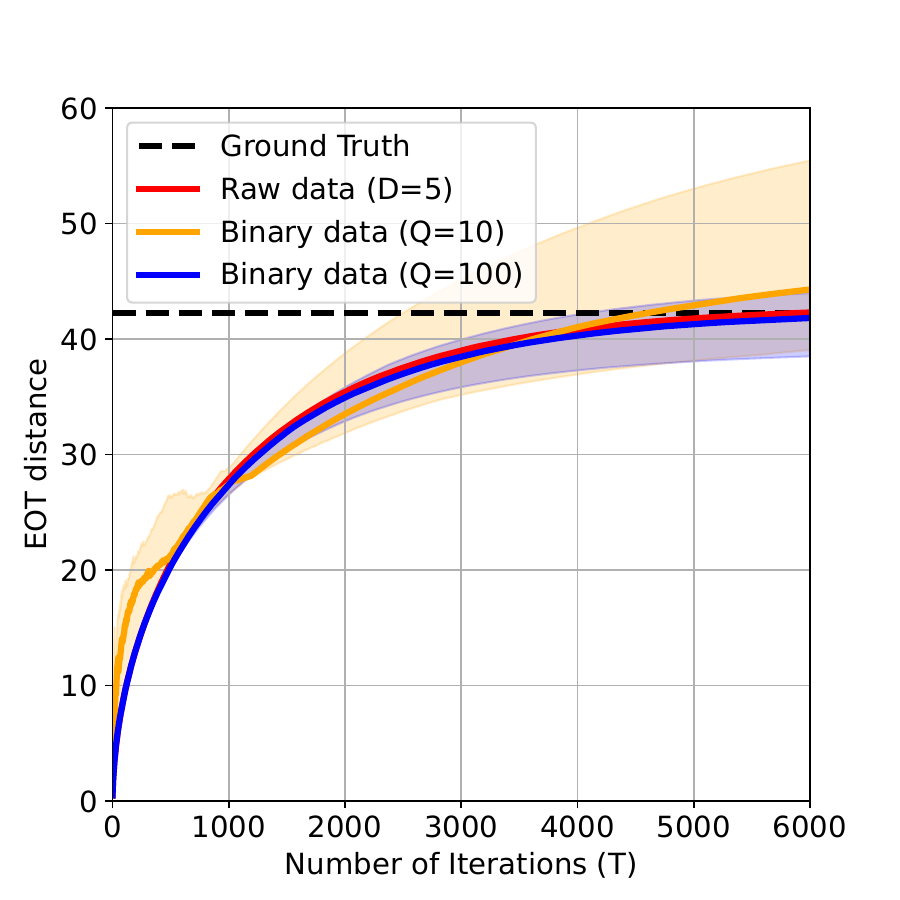}
    }
    \subfigure[$W_{\varepsilon}(\mathcal{N}_1,\mathcal{N}_2)$, $L=8$]{
    \includegraphics[height=4.0cm]{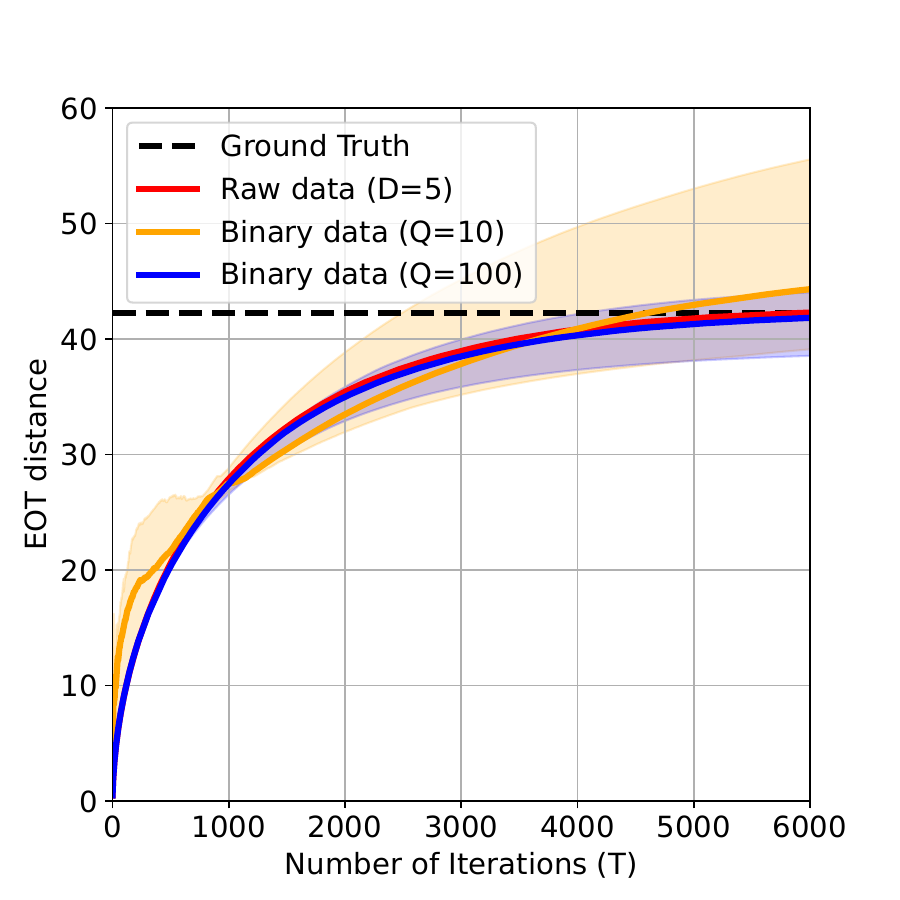}
    }
    \subfigure[$W_{\varepsilon}(\mathcal{M}_1,\mathcal{M}_2)$, $L=1$]{
    \includegraphics[height=4.0cm]{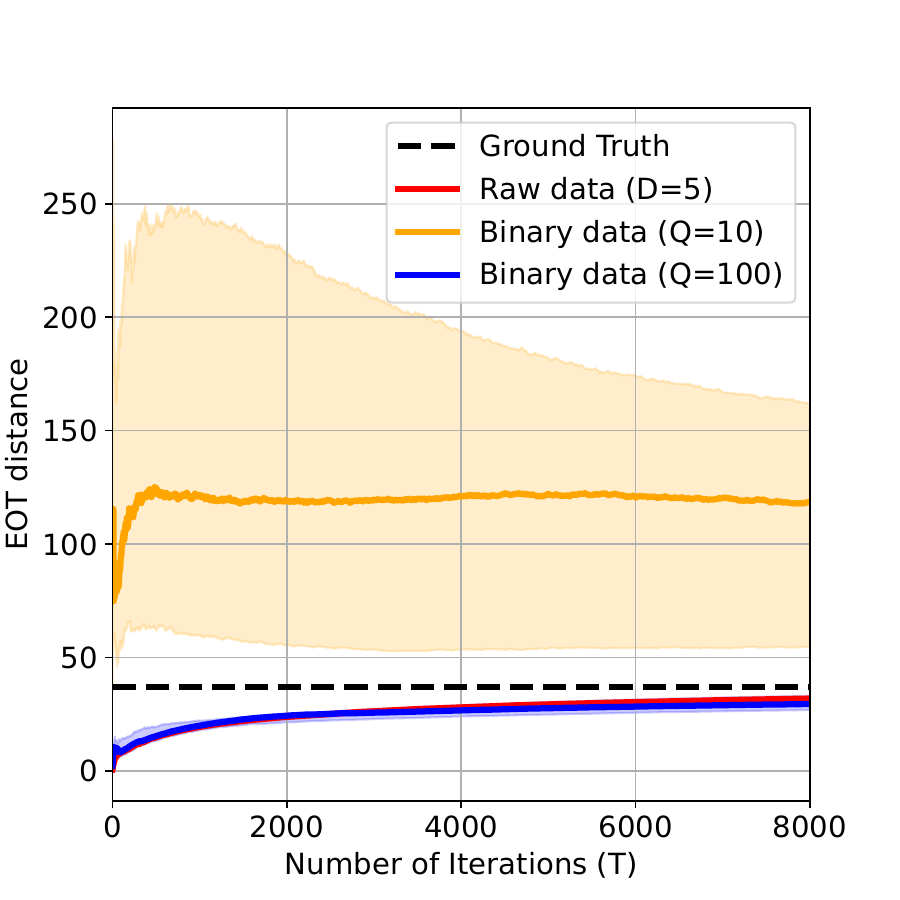}
    }
    \subfigure[$W_{\varepsilon}(\mathcal{M}_1,\mathcal{M}_2)$, $L=4$]{
    \includegraphics[height=4.0cm]{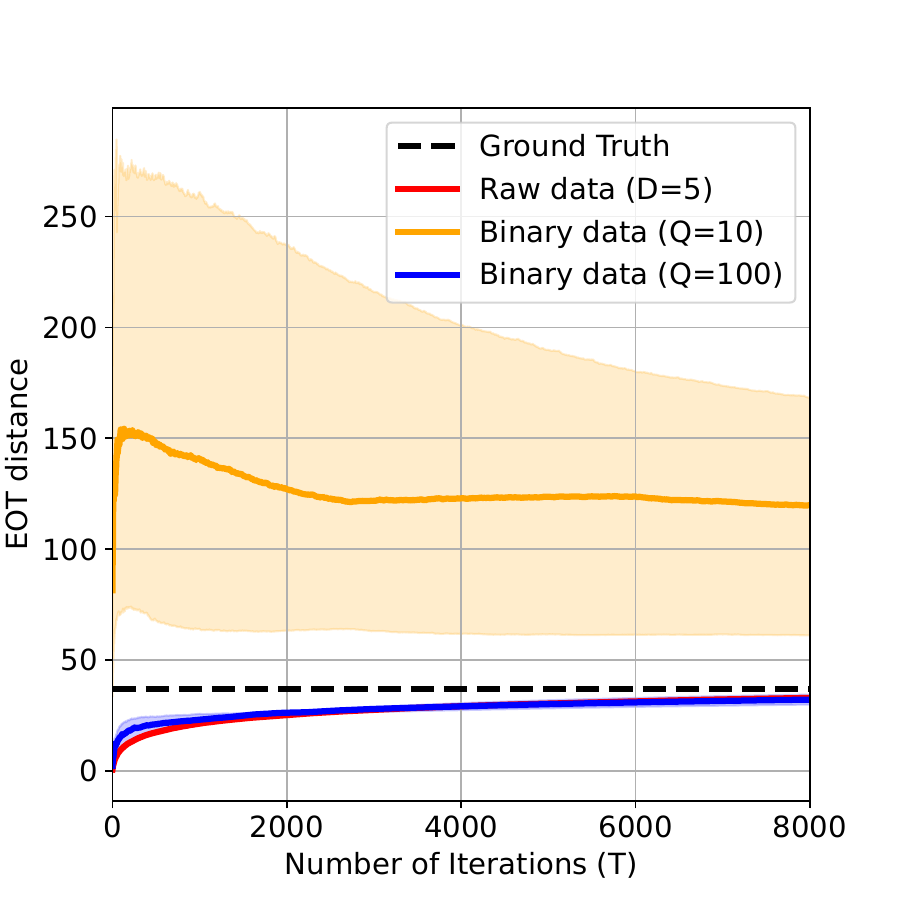}
    }
    \subfigure[$W_{\varepsilon}(\mathcal{M}_1,\mathcal{M}_2)$, $L=8$]{
    \includegraphics[height=4.0cm]{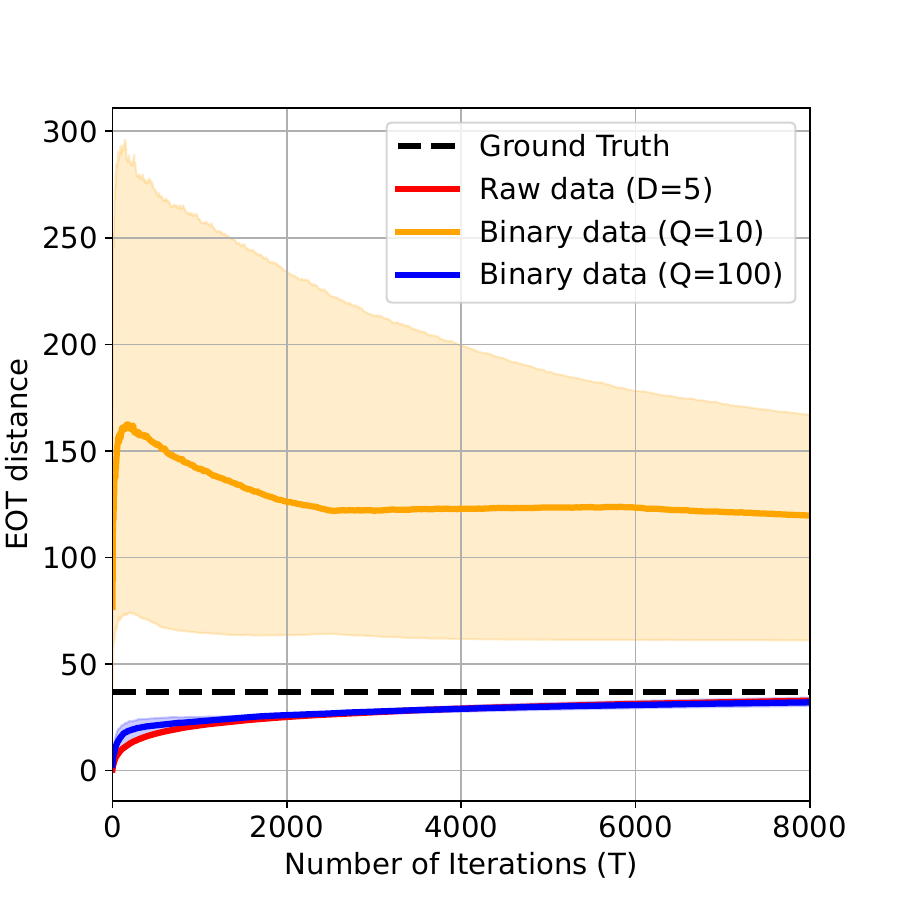}
    }
    \subfigure[$W_{\varepsilon}(\mathcal{M}_1,\mathcal{M}_2)$, $L=1$]{
    \includegraphics[height=4.0cm]{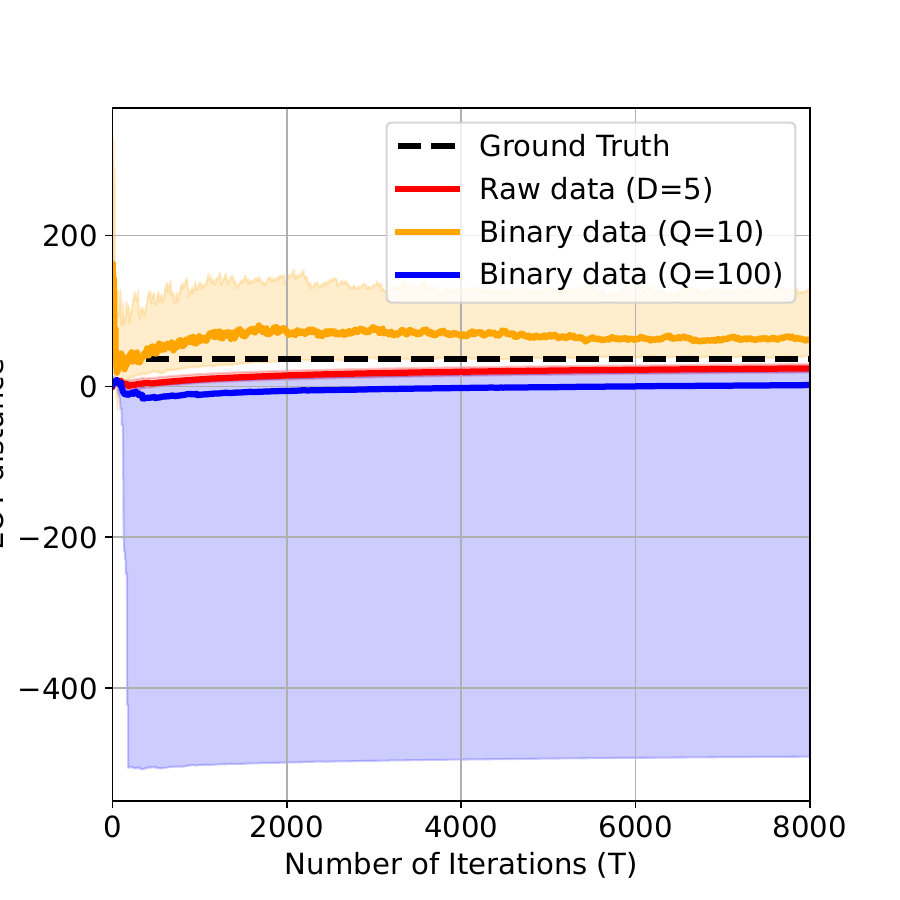}
    }
    \subfigure[$W_{\varepsilon}(\mathcal{M}_1,\mathcal{M}_2)$, $L=4$]{
    \includegraphics[height=4.0cm]{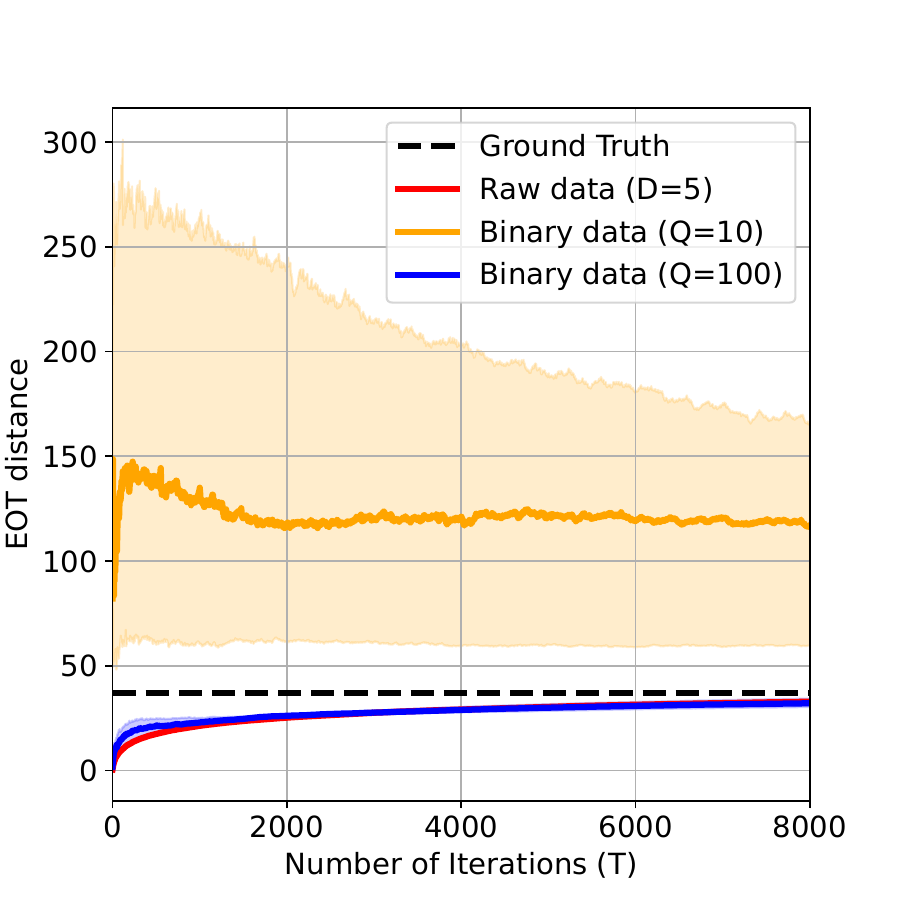}
    }
    \subfigure[$W_{\varepsilon}(\mathcal{M}_1,\mathcal{M}_2)$, $L=8$]{
    \includegraphics[height=4.0cm]{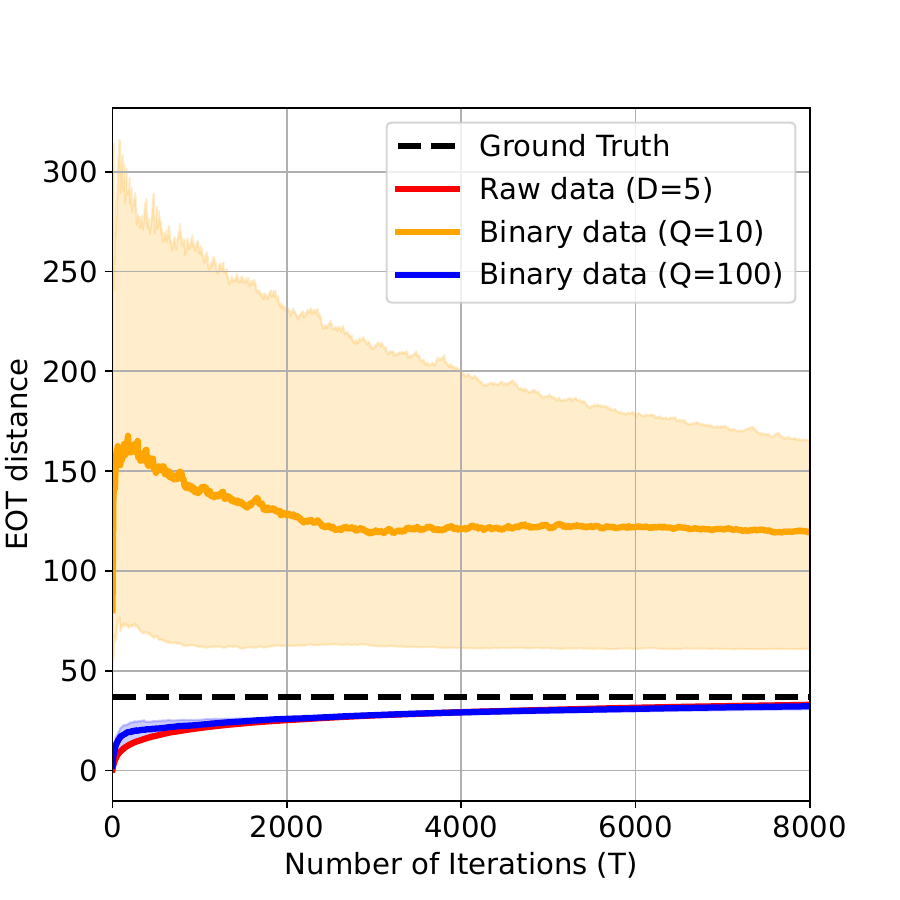}
    }
    \caption{In each subfigure, the block dotted line indicates the EOT computed by the classic Sinkhorn-scaling algorithm. 
    The red, orange, and blue curves indicates the average convergence curves of our DEOT method when applying raw data or binary data, respectively. 
    (a-c) show the results of computing $W_{\varepsilon}(\mathcal{N}_1,\mathcal{N}_2)$.
    (d-f) show the results of computing $W_{\varepsilon}(\mathcal{M}_1,\mathcal{M}_2)$ in the i.i.d. setting.
    (g-i) show the results of computing $W_{\varepsilon}(\mathcal{M}_1,\mathcal{M}_2)$ in the non-i.i.d. setting.}
    \label{fig:eot_kernel}
\end{figure}

\subsubsection{Robustness to $L$}
We approximate $W_{\varepsilon}(\mathcal{N}_1,\mathcal{N}_2)$ and $W_{\varepsilon}(\mathcal{M}_1,\mathcal{M}_2)$ via our DEOT method, in which $L\in\{1, 4, 8\}$ and the kernel matrix can be the $\bm{K}$ based on raw data or the $\widehat{\bm{K}}$ based on binary vectors (with $Q\in\{10, 100\}$).
We compare the results with the ground truth achieved by the centralized Sinkhorn-scaling algorithm~\cite{cuturi2013sinkhorn}. 
Fig.~\ref{fig:eot_kernel} visualizes the convergence of our method under different settings.
Our method is robust to $L$ when the samples are randomly scattered to different agents. 
As shown in Fig.~\ref{fig:eot_kernel}(a-f), even if we set $L=1$ (i.e., only consider one agent when computing the gradients in each iteration), the performance of our method is comparable to that achieved when setting $L=4$ or $L=8$.  
However, when the samples of different agents are non-i.i.d., the gradients computed based on one agent are biased and thus cause undesired performance, as shown in Fig.~\ref{fig:eot_kernel}(g). 
In other words, in non-i.i.d. scenarios, we need to consider more agents when computing gradients.

\subsubsection{The Impact of Kernel Approximation}

In Fig.~\ref{fig:eot_kernel}, we also compare the performance of our DEOT method when communicating raw data to that when communicating binary vectors. 
We can find that when the dimension of the binary vector (i.e., $Q$) is high enough, e.g., $Q=100$, the error and the variance caused by kernel approximation are tolerable. 
As we show in~\eqref{eq:error_kernel_approx}, this error is independent of the dimension of raw data but linear with the number of samples and $Q^{-1}$. 
Therefore, the more samples we have, the higher dimension $Q$ we need. 

\begin{figure}[t]
    \centering
    \includegraphics[height=6cm]{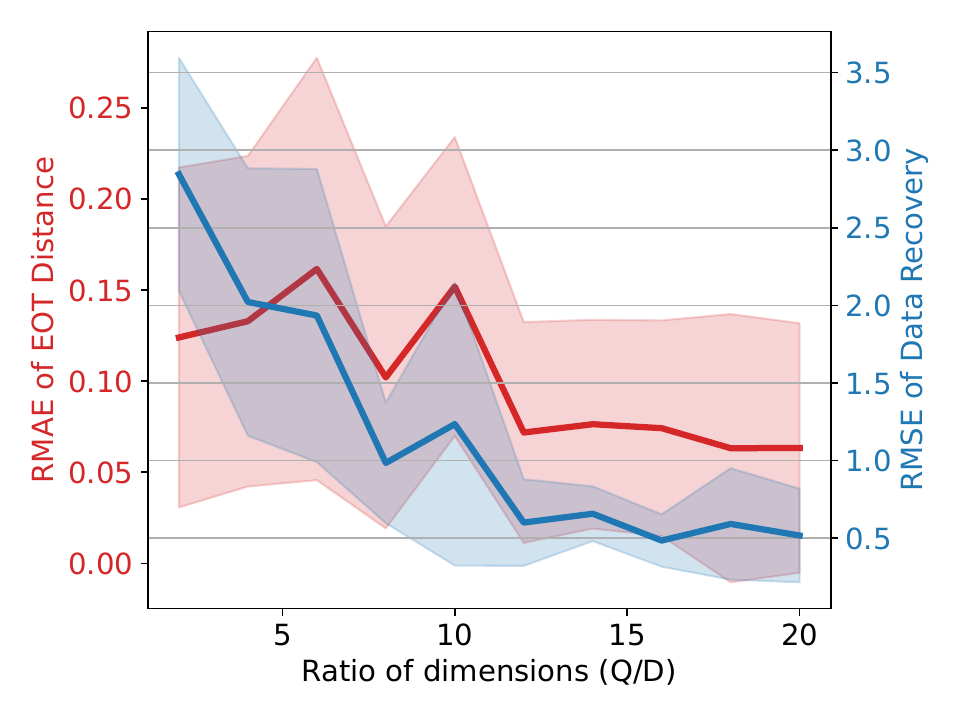}
    \caption{The RMAE of EOT distance and the RMSE of sample estimation with respect to $\frac{Q}{D}$.}
    \label{fig:privacy}
\end{figure}

Note that the higher dimension the binary vectors have, the more information is shared during communication. 
In particular, for each agent with some samples, it may receive the norm and the binary vector of a sample from the other agent and, accordingly, approximate the kernel-based similarity (and equivalently, the distance) between this sample and each of its own samples. 
As long as the number of its samples $N$ is larger than the sample dimension $D$, the agent can likely estimate the received sample using least-square estimation. 
In Fig.~\ref{fig:privacy}, we apply our DEOT method with different $Q$'s and consider two evaluation metrics: $i)$ the RMAE of the EOT distance, i.e., $\frac{|F_{\varepsilon}(\hat{\bm{u}}^t,\hat{\bm{v}}^t;\widehat{\bm{K}},\bm{E})-W_{\varepsilon}(\mu,\gamma)|}{W_{\varepsilon}(\mu,\gamma)}$, and $ii)$ the RMSE of data recovery, i.e., $\frac{\|\hat{\bm{y}}-\bm{y}\|_2}{\|\bm{y}\|_2}$, where $\bm{y}$ is a sample of a target agent and $\hat{\bm{y}}$ is the least-square estimation achieved by a source agent after the agent obtained the norm and the binary vector of $\bm{y}$.
The former measures the precision of our DEOT method when computing EOT distance, while the latter measures the strength of data privacy protection. 
As shown in Fig.~\ref{fig:privacy}, with the increase of $Q$, both two metrics reduce. 
Fortunately, when the RMAE of EOT distance is significantly small (e.g., $\sim 0.05$), the RMSE of recovered data is still larger than $0.5$.
In other words, in practice, we can set $Q$ robustly in a wide range (e.g., $\frac{Q}{D}>15$) to achieve a trade-off between the precision of our method and the strength of data privacy protection.

\begin{figure}[t]
    \centering
    \subfigure[$W_{\varepsilon}(\mathcal{N}_1,\mathcal{N}_2)$, i.i.d.]{
    \includegraphics[height=4.0cm]{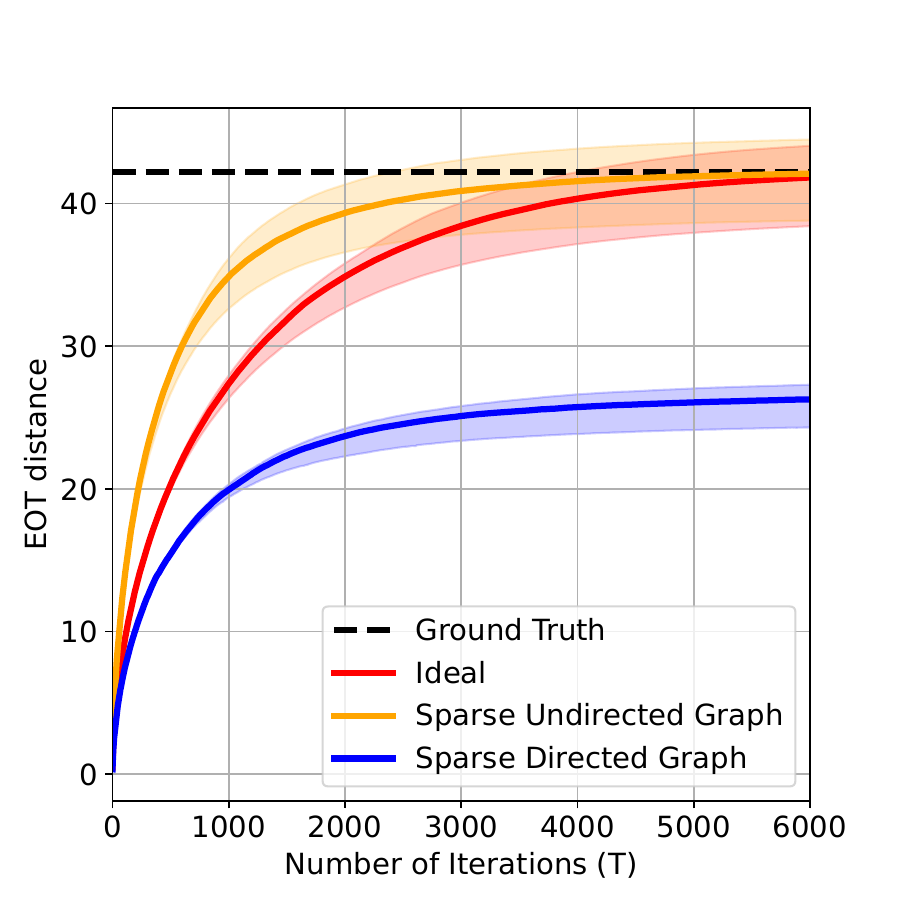}
    }
    \subfigure[$W_{\varepsilon}(\mathcal{M}_1,\mathcal{M}_2)$, i.i.d.]{
    \includegraphics[height=4.0cm]{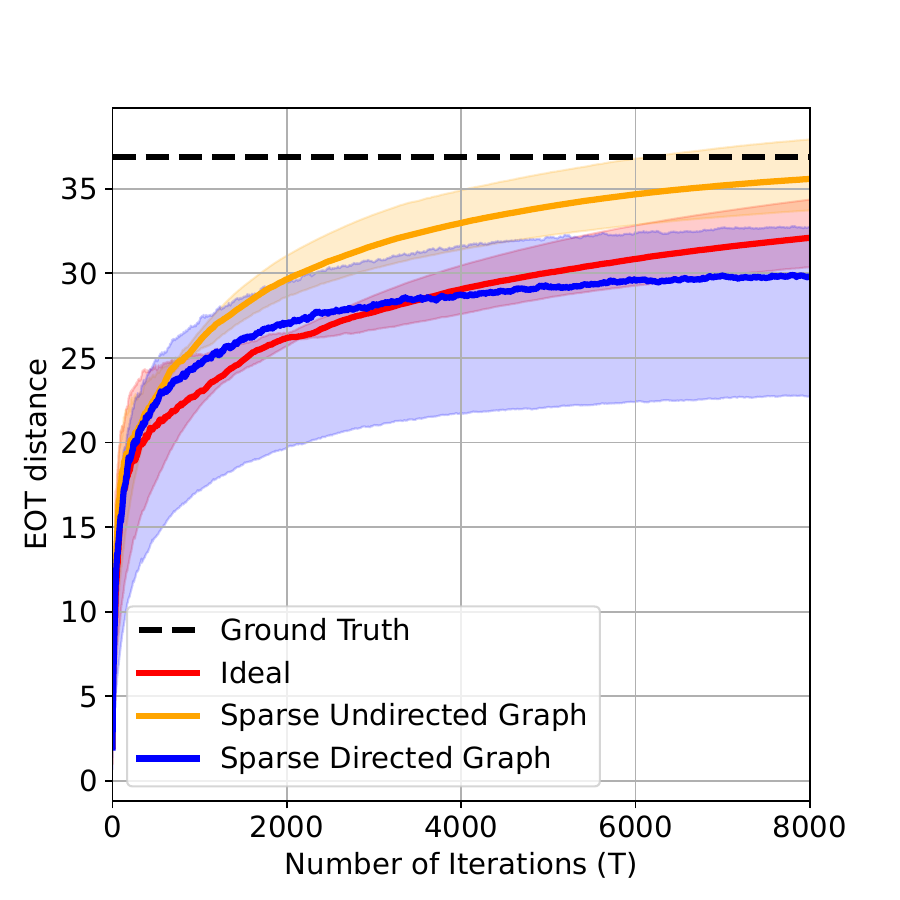}
    }
    \subfigure[$W_{\varepsilon}(\mathcal{M}_1,\mathcal{M}_2)$, non-i.i.d.]{
    \includegraphics[height=4.0cm]{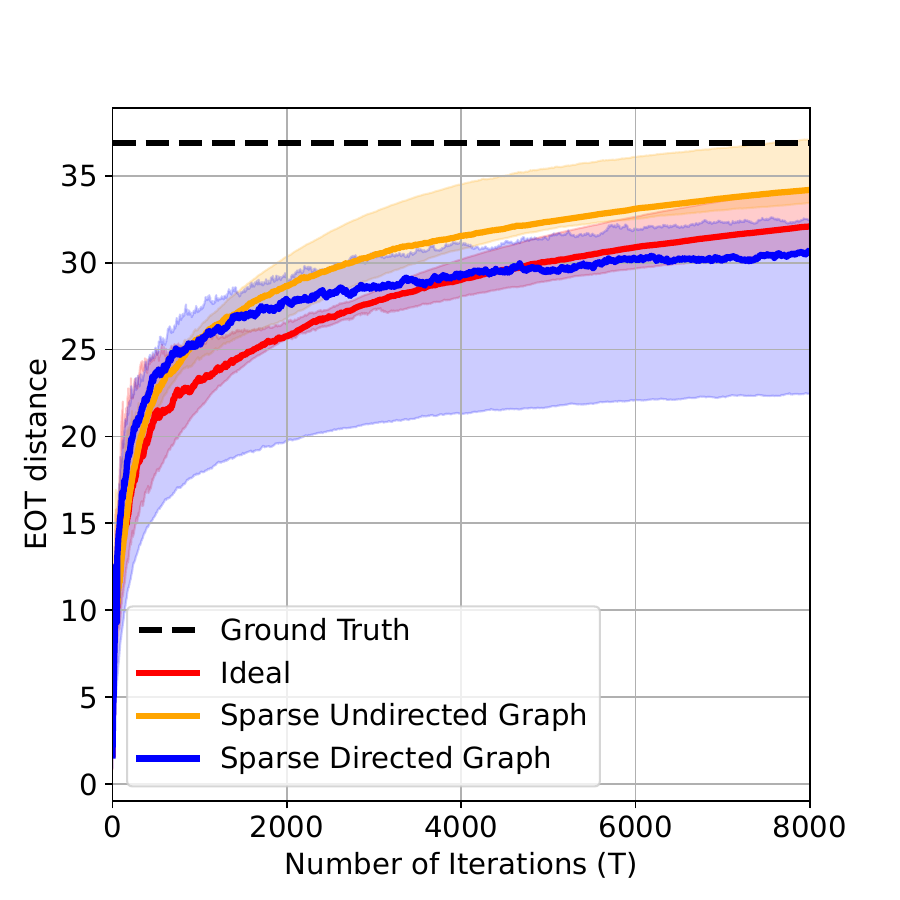}
    }
    \caption{In each subfigure, the block dotted line indicates the EOT computed by the classic Sinkhorn-scaling algorithm. 
    The red, orange, and blue curves indicates the average convergence curves of our DEOT method under different communication protocols.}
    \label{fig:eot_protocol}
\end{figure}

\subsubsection{The Impact of Communication Protocol} 

As shown in Lemma~\ref{lm:protocol}, the communication protocol impacts our DEOT method significantly. 
In Fig.~\ref{fig:eot_protocol}, we apply our method with three different communication protocols: $i)$ the ideal communication protocol perfectly matching with the storage protocol, i.e., $\bm{E}=\bm{pq}^{\top}=[\frac{1}{IJ}]$, $ii)$ a sparse $\bm{E}$ defined on a sparse undirected graph, e.g., each source agent only communicate with four target agents (50\% zeros in $\bm{E}$), and $iii)$ a sparse and asymmetric $\bm{E}$ defined on a sparse directed graph, e.g., setting the upper-triangular part of the sparse $\bm{E}$ to be all-zero (directed communication).
Note that the sparse undirected graph in this experiment still corresponds to a connected network, while the connectivity of the sparse directed graph is not guaranteed.

Experimental results in Fig.~\ref{fig:eot_protocol} show that the results corresponding to the sparse undirected graph are comparable to those in the ideal scenarios. 
In other words, as long as the connectivity is guaranteed, the errors caused by mismatched protocols are tolerable.
On the contrary, the deterioration of the communication environment leads to significant performance degradation --- when the communications happen in a sparse directed graph, whose connectivity is not guaranteed, the estimated EOT distance either is far from the ground truth or suffers high variance.

\begin{figure}[t]
    \centering
    \subfigure[$GW_{\varepsilon}(\mathcal{N}_1,\mathcal{N}_2)$, $L=1$]{
    \includegraphics[height=4.0cm]{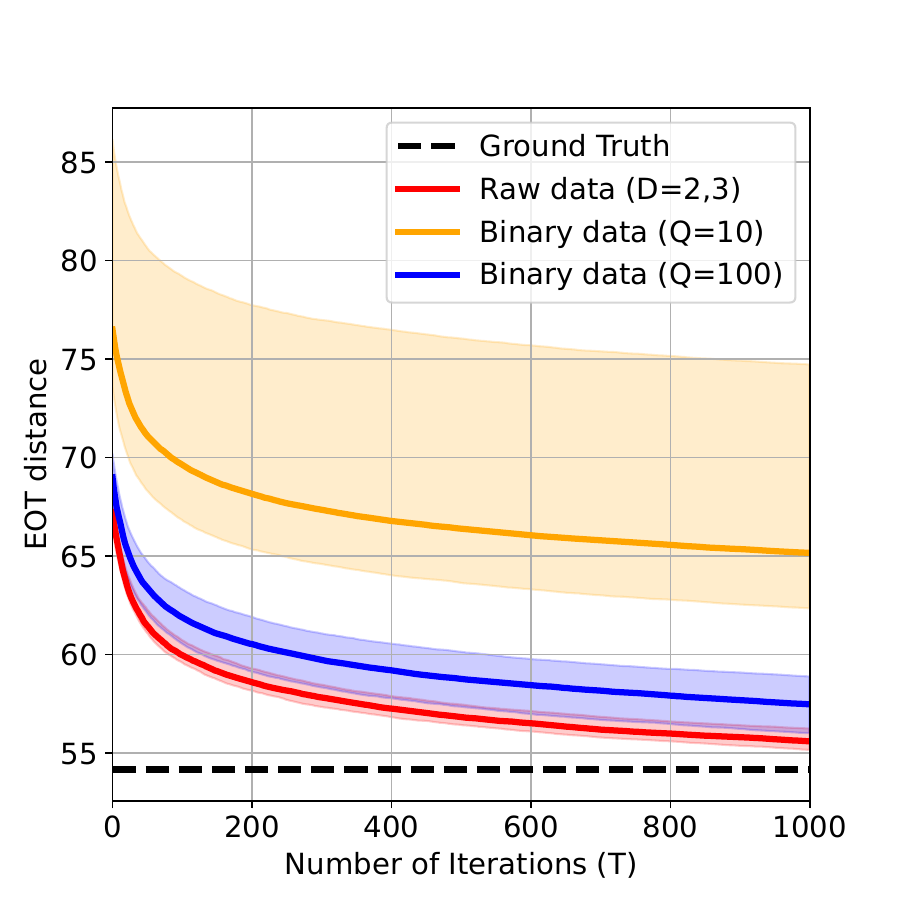}
    }
    \subfigure[$GW_{\varepsilon}(\mathcal{N}_1,\mathcal{N}_2)$, $L=4$]{
    \includegraphics[height=4.0cm]{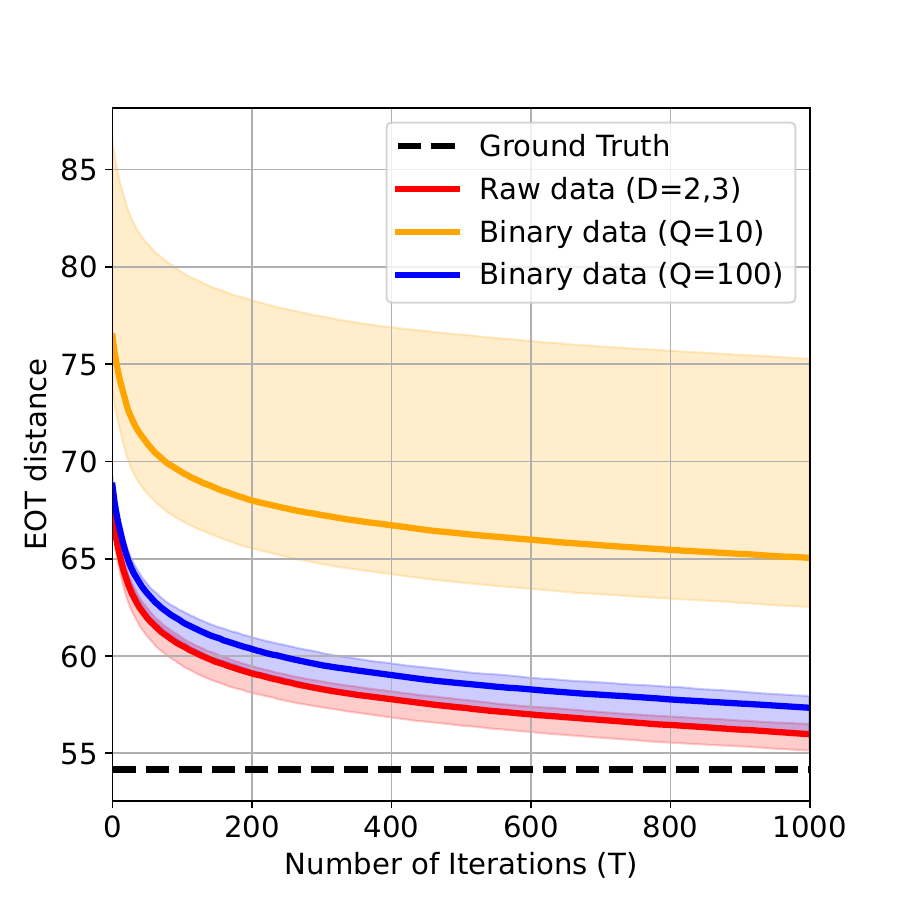}
    }
    \subfigure[$GW_{\varepsilon}(\mathcal{N}_1,\mathcal{N}_2)$, $L=8$]{
    \includegraphics[height=4.0cm]{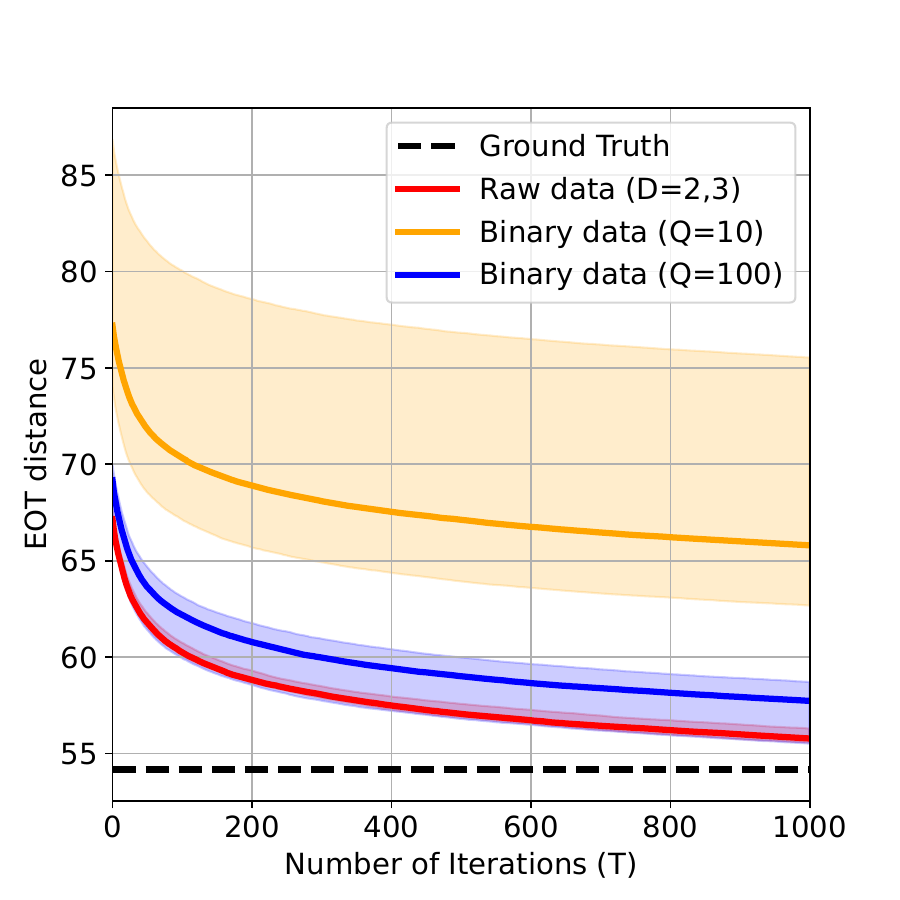}
    }
    \caption{In each subfigure, the block dotted line indicates the EGW computed by the classic centralized proximal gradient algorithm~\cite{peyre2016gromov}. 
    The red, orange, and blue curves indicates the average convergence curves of our DEGW method under different communication protocols.}
    \label{fig:egw_kernel}
\end{figure}

\subsubsection{Effectiveness on Computing EGW Distance}
As aforementioned, we can extend our DEOT method to a DEGW method when computing EGW distance. 
To demonstrate the effectiveness of our method, 
we consider the EGW distance between a 2D Gaussian distribution and a 3D Gaussian distribution.
Given the two distributions, we generate 2,000 samples from each and scatter their samples randomly to a distributed system with eight source agents and eight target agents. 
Taking the EGW distance obtained by the classic centralized proximal gradient algorithm~\cite{peyre2016gromov} as the ground truth, we test the proposed DEGW method (i.e., Algorithm~\ref{Algo-MRBCD2}) and check whether its result can approach the ground truth or not. 
Experimental results in Fig.~\ref{fig:egw_kernel} show that with the increase of iteration, our DEGW method can approximate the EGW distance well, whose results converge to the ground truth.
Similar to the results in Fig.~\ref{fig:eot_kernel}, our method is robust to $L$ --- in the i.i.d. scenario, setting $L=1$ can still achieve encouraging performance.
In addition, with the increase of $Q$, the error caused by the approximate kernel is mitigated, and the convergence curve of our DEGW method becomes close to that corresponding to using raw data.

\subsection{Real-world distributed domain adaptation}
Besides the synthetic experiments above, we conduct real-world distributed domain adaptation experiments in classification tasks. 
In particular, distributed domain adaptation is dedicated to solving the domain adaptation problem where both source and target domain data are scattered over different agents. The aim is to use the label information available in the source domain $\mathcal{X}$ to learn a classifier that can be applied to the target domain $\mathcal{Y}$ without label information.
Unlike classic domain adaptation, the distributed domain adaptation needs to consider the communications among different agents. 
Suppose we further protect the privacy of the target domain by preventing the source domain from accessing the target domain's data. 
The problem is even more challenging, and existing domain adaptation methods become inapplicable. 

We focus on the OT-based domain adaptation strategy~\cite{flamary2016optimal}.
This strategy $i)$ computes the (entropic) optimal transport distance between the source and target domains, $ii)$ maps the source samples to the target domain via the optimal coupling, and $iii)$ trains the 1-Nearest Neighbor (1NN) classifier based on the mapped data. 
To obtain the optimal coupling, we apply various methods, including our DEOT method with real or approximated kernel matrix, the earth mover distance (EMD) for OT distance, the Sinkhorn algorithm for EOT distance, and the OT-LpL1 method in~\cite{courty2014domain}. 
The baselines (EMD, Sinkhorn, OT-LpL1) are centralized and designed for classic domain adaptation.
For a fair comparison and to highlight our contribution, we test the baselines under their default centralized settings in the following experiments while testing our DEOT method under the decentralized setting.

\subsubsection{Implementation Details}
Our DEOT method provides a promising solution to distributed domain adaptation. 
Specifically, suppose we have the source domain data $\bm{X}_{i}=\{\bm{x}_{n}^{(i)}\}_{n=1}^{N_i}$ associated with the class labels, and the target domain data $\bm{Y}_{j}=\{\bm{y}_{m}^{(j)}\}_{m=1}^{M_j}$ with unknown labels. 
Based on our DEOT method, each target agent $j$ can obtain an optimal coupling $\{\bm{\Pi}_{ij}=\text{diag}(\hat{\bm{u}}^{(i),T})\bm{K}_{ij}\text{diag}(\hat{\bm{v}}^{(j),T})\}_{i=1}^{I}$, where $\hat{\bm{u}}^{(i),T}$ and $\hat{\bm{v}}^{(j),T}$ are optimized dual variables after $T$-step updating. 
Then, according to~\cite{courty2017joint}, when the probability measures $\mu$ and $\gamma$ are uniform, we can derive the barycentric mapping as $\widehat{\bm{X}}=N \bm{\Pi} \bm{Y}$, where $\bm{\Pi}=[\bm{\Pi}_{ij}]$ is the complete coupling and $\widehat{\bm{X}}$ is the transported data of the source domain. 
In our setting, this barycentric mapping can be achieved in a decentralized way.
In particular, the source agent $i$ first receives the dual variables of target agents and computes $\bm{\Pi}_{ij}$'s. 
Then, it can send $\bm{\Pi}_{ij}$'s to the corresponding target agents and receive $\bm{\Pi}_{ij}\bm{Y}_j$ accordingly.
The aggregation of the received data, i.e., $\sum_{j}\bm{\Pi}_{ij}\bm{Y}_j$, leads to the transported data $\widehat{\bm{X}}_i$ of the source agent $i$. 
Eventually, we can train the 1NN classifier given the transported data $\widehat{\bm{X}}$ and perform classification prediction on the target domain data.
Note that each source agent can only receive $\bm{\Pi}_{ij}\bm{Y}_j$ rather than raw data, and our DEOT method can compute $\bm{\Pi}_{ij}$ without sharing raw data, so the distributed domain adaptation achieved by our method can be privacy-preserving to some extent.

\begin{table}[t]
\centering
\caption{Summary of the domains used in the experiments}\label{tab:sum}
\begin{small}
\begin{tabular}{@{}c|c|c|c|c|c@{}}
\hline\hline
Problem   
& Domains   
& Datasets 
& \#Samples
& \#Features 
& Abbr.\\
\hline

\multirow{2}{*}{Digits} & USPS           & USPS   & $1,800$ & $256$  & U\\

  & MNIST      & MNIST   & $2,000$  & $256$  & M\\
 \hline
\multirow{4}{*}{Objects} & Art & Office-home & $2,427$ & $2,048$ & Ar \\
& Clipart & Office-home & $4,365$ & $2,048$ & Cl \\
& Product & Office-home & $4,439$ & $2,048$ & Pr \\
& Real-World & Office-home & $4,357$ & $2,048$ & Rw \\
\hline\hline
\end{tabular}
\end{small}
\end{table}

We conduct this experiment on two widely-used domain adaptation datasets. 
The first is the digital number adaptation dataset of USPS and MNIST~\cite{lecun1998gradient}. 
For USPS and MNIST, each has ten image categories corresponding to the digits from 0 to 9.
We follow the setting in~\cite{courty2017joint}. 
Given 2,000 images of the MNIST domain and 1,800 images of the USPS domain, we consider the adaptation in two directions: \textbf{U}SPS$\rightarrow$\textbf{M}NIST and \textbf{M}NIST$\rightarrow$\textbf{U}SPS. 
The second is the Office-home dataset~\cite{2017Deep}.
The Office-home dataset contains around 15,500 images in four different domains: Art (artistic images in the form of sketches, paintings, and so on), Clipart (a collection of clipart images), Product (images of objects without a background), and RealWorld (images of objects captured with a regular camera). 
Based on this dataset, we consider 12 transfer tasks for the Art (\textbf{Ar}), Clipart (\textbf{Cl}), Product (\textbf{Pr}) and Real-World (\textbf{Rw}) domains for all combinations of source and target for the four domains. 
A summary of the properties of each domain used in this paper is provided in Table~\ref{tab:sum}.

For the experimental setup, we scattered the source and target domain data over four agents and set $L=7$ and $Q=10,000$.
The samples of each domain are features extracted through a pre-trained ResNet-50~\cite{DBLP:journals/corr/HeZRS15}. 
For our DEOT method, we apply grid search, finding the optimal weight of regularizer $\varepsilon\in \{2, 1, 0.5, 0.1, 0.05\}$ and the optimal learning rate $\eta\in\{1, 0.5, 0.1, 0.01, 0.001\}$.

\subsubsection{Experimental results}
Experimental results in Table~\ref{tab:officehome} show that without the information of the target domain, purely training a 1NN classifier leads to unsatisfactory performance. 
The traditional centralized OT methods can improve classification accuracy. 
Still, they require a powerful central server to compute the OT distance and need to access the raw data of the target domain.
Our method outperforms the baselines when using the real kernel and achieves privacy preservation with tolerable performance degradation when using the approximated kernel. 
In summary, our DEOT method has the potential for these distributed domain adaptation tasks, especially in challenging privacy-preserving scenarios.


\begin{table}[t]
\centering
\caption{Comparisons on classification accuracy in distributed domain adaptation tasks}\label{tab:officehome}
\begin{small}
\begin{tabular}{@{}c|c|ccc|cc@{}}
\hline\hline
\multirow{2}{*}{Domains}     
& Source only
& 
&Centralized
&
&
\multicolumn{2}{c}{Decentralized (Ours)} \\ 

& 1NN
& EMD
& Sinkhorn
& OT-LpL1

& DEOT$_{\bm{K}}$
& DEOT$_{\widehat{\bm{K}}}$\\
\hline
U$\rightarrow$M    & $0.385$   & $0.554$  & $0.437$  & $0.490$ & $0.580$ & $0.522$\\ 
M$\rightarrow$U    & $0.593$   & $0.617$  & $0.620$  & $0.676$ & $0.681$ & $0.629$\\
\hline
Ar$\rightarrow$Cl    & $0.433$   & $0.471$  & $0.492$  & $0.490$ & $0.483$ & $0.458$\\ 
Ar$\rightarrow$Pr    & $0.594$   & $0.642$  & $0.673$  & $0.633$ & $0.665$ & $0.639$\\
Ar$\rightarrow$Rw    & $0.667$   & $0.677$  & $0.721$  & $0.686$ & $0.738$ & $0.705$\\
\hline
Cl$\rightarrow$Ar   & $0.445$   & $0.504$  & $0.509$  & $0.478$ & $0.531$ & $0.509$\\
Cl$\rightarrow$Pr    & $0.536$   & $0.647$  & $0.617$  & $0.642$ & $0.632$ & $0.606$\\
Cl$\rightarrow$Rw    & $0.589$   & $0.638$  & $0.657$  & $0.664$ & $0.654$ & $0.618$\\
\hline
Pr$\rightarrow$Ar    & $0.488$   & $0.516$  & $0.532$  & $0.494$ & $0.538$ & $0.506$\\
Pr$\rightarrow$Cl    & $0.414$   & $0.455$  & $0.465$  & $0.450$ & $0.469$ & $0.425$\\
Pr$\rightarrow$Rw    & $0.683$   & $0.707$  & $0.725$  & $0.714$ & $0.735$ & $0.704$\\
\hline
Rw$\rightarrow$Ar    & $0.592$   & $0.611$  & $0.622$  & $0.605$ & $0.621$ & $0.598$\\
Rw$\rightarrow$Cl    & $0.450$   & $0.498$  & $0.505$  & $0.509$ & $0.494$ & $0.463$\\
Rw$\rightarrow$Pr    & $0.729$   & $0.749$  & $0.778$  & $0.770$ & $0.773$ & $0.736$\\

\hline\hline
\end{tabular}
\end{small}
\end{table}